\documentclass[12pt,a4paper,notitlepage]{article}

\usepackage[english]{babel}
\usepackage[utf8]{inputenc}
\usepackage[T1]{fontenc}
\usepackage{lmodern}

\usepackage{tipa}  
\usepackage{CJKutf8}
\usepackage{newunicodechar}
\DeclareUnicodeCharacter{202F}{\,}  
\usepackage{xcolor}
\definecolor{mauve}{rgb}{0.58,0,0.82}
\usepackage{sectsty}
\sectionfont{\color{blue}}
\subsectionfont{\color{red}}

\usepackage{amsmath,amssymb,amsfonts}
\usepackage{mathtools}
\usepackage{systeme}
\usepackage{siunitx}

\usepackage{multirow}
\usepackage{makecell}
\usepackage{booktabs}
\usepackage{colortbl}
\usepackage[table]{xcolor}
\usepackage{ltablex,ragged2e}
\usepackage{array}
\usepackage{blkarray}
\newcolumntype{L}[1]{>{\raggedright\arraybackslash}p{#1}}

\usepackage{graphicx}
\usepackage{caption}
\usepackage{pgfplotstable}
\pgfplotsset{compat=1.8}

\usepackage[normalem]{ulem}
\usepackage{soul}
\usepackage{etoolbox}
\usepackage{enumitem,lipsum}
\usepackage{setspace}

\usepackage{lineno}

\usepackage{tcolorbox}
\usepackage[colorinlistoftodos]{todonotes}

\usepackage{url}
\usepackage{amssymb}
\usepackage{pifont}

\usepackage{authblk}

\usepackage{hyperref}
\hypersetup{
	colorlinks=true,
	linkcolor=blue,
	citecolor=blue,
	urlcolor=red
}

\date{\vspace{-5ex}}

\title{Systematic Framework of Application Methods for Large Language Models in Language Sciences}

\author[1,2]{\footnotesize Kun Sun\thanks{Corresponding author: \texttt{kunsun@tongji.edu.cn}}}
\author[2,3]{\footnotesize Rong Wang\thanks{Corresponding author: \texttt{rong.wang@uni-tuebingen.de}}}
\affil[1]{\footnotesize Tongji University, China}
\affil[2]{\footnotesize University of Tübingen, Germany}
\affil[3]{\footnotesize Institute of Natural Language Processing, University of Stuttgart, Germany}

\begin{document}

\maketitle


	\begin{abstract}
	\linespread{1.0}\selectfont
	Large Language Models (LLMs) are transforming language sciences. However, their widespread deployment currently suffers from methodological fragmentation and a lack of systematic soundness. This study proposes two comprehensive methodological frameworks designed to guide the strategic and responsible application of LLMs in language sciences. The first method-selection framework defines and systematizes three distinct, complementary approaches, each linked to a specific research goal: (1) prompt-based interaction with general-use models for exploratory analysis and hypothesis generation; (2) fine-tuning of open-source models for confirmatory, theory-driven investigation and high-quality data generation; and (3) extraction of contextualized embeddings for further quantitative analysis and probing of model internal mechanisms. We detail the technical implementation and inherent trade-offs of each method, supported by empirical case studies. Based on the method-selection framework, the second systematic framework proposed provides \textit{constructed configurations} that guide the practical implementation of multi-stage research pipelines based on these approaches. We then conducted a series of empirical experiments to validate our proposed framework, employing retrospective analysis, prospective application, and an expert evaluation survey. By enforcing the strategic alignment of research questions with the appropriate LLM methodology, the frameworks enable a critical paradigm shift in language science research. We believe that this system is fundamental for ensuring reproducibility, facilitating the critical evaluation of LLM mechanisms, and providing the structure necessary to move traditional linguistics from ad-hoc utility to verifiable, robust science.
\end{abstract}






\section{Introduction}

Large language models (LLMs) built on the Transformer architecture have demonstrated impressive performance on a wide range of language processing tasks \cite{devlin2019,brown2020}. Models such as \texttt{BERT} family and \texttt{GPT}-series, trained on billions of parameters and massive text corpora, enable applications in language understanding, generation, and analysis \cite{chang2024, ambridge2024, begus2023, cai2025}. Their exceptional linguistic capabilities have drawn growing attention from researchers in the language sciences who seek computational tools to extend and enrich traditional methodologies.

Language sciences, as discussed here, extend beyond traditional linguistics by integrating computational, psychological, and neuroscience approaches to study how language is represented, processed, and learned. It encompasses a wide range of subfields, including corpus linguistics, historical linguistics, psycholinguistics, and experimental linguistics. Unlike traditional linguistics, which often emphasizes formal description and theoretical modeling, language science prioritizes empirical validation and quantitative methods to understand language as both a cognitive and communicative system.

LLMs have been widely and intensively applied in language science research. They can efficiently process large-scale corpora, enabling analyses that would be infeasible using manual methods. LLMs provide quantitative metrics for linguistic phenomena that previously relied on subjective judgment and can generate stimuli for experimental studies or assist with annotation tasks that require substantial human effort \cite{weissweiler2024hybrid, barbieri2020, movva2024annotation, ziems2024can,lang2025using}. At the same time, their rapid adoption has highlighted several methodological and conceptual challenges that must be addressed to fully realize their potential in the study of language.

First, the current landscape of LLM applications in language science is characterized by methodological atomization and a reliance on task-specific heuristics rather than a unified, scientific framework, that is, methodological fragmentation is pervasive. Studies often adopt different models, hyperparameters, metrics, and reporting standards, which impedes comparability \cite{katzir2023}. Existing research frequently exhibits a 'method-as-end' tendency, in which researchers choose an LLM application paradigm, such as simple prompting or blind fine-tuning, based on familiarity or model availability rather than on clear, hypothesis-driven objectives.  This fragmentation creates two core challenges. First, there is often a mismatch between research goals and methods. For example, theoretical work aimed at probing a model's internal linguistic representations is sometimes approached using black-box prompting, which is appropriate only for exploratory data generation. Second, this lack of systemic soundess breeds concerns over replicability, external validity, and the cumulative nature of knowledge derived from LLM experiments, hindering the field's transition to mature science. Linguistic researchers are left navigating a disjointed decision space, often resorting to trial-and-error in selecting the most appropriate LLM application technique (prompting or probing) for their specific theoretical goals. 

Second, preprocessing steps and parameter choices are often underreported, limiting reproducibility, and model selection is frequently made without clear justification relative to the research question, raising concerns about interpretability and theoretical alignment \cite{peng2025recalibrating}. Tokenization, embedding extraction, and filtering procedures can all influence outcomes \cite{bai2024beyond}. However, these details are rarely fully documented. Such issues highlight the need for transparent reporting and careful method alignment. Ultimately, addressing these challenges requires a unified framework to guide model selection, preprocessing, and analysis, ensuring rigorous, interpretable, and reproducible LLM research in the language sciences.

Third, conceptual ambiguity remains regarding what model performance reveals about language. High task accuracy alone does not validate theoretical claims, as multiple mechanisms may produce similar outputs. Conversely, model failures do not necessarily disprove linguistic theories, since implementation details can influence performance. Different subdisciplines maintain distinct evidentiary standards and theoretical commitments, which must be respected when interpreting LLM results \cite{katzir2023,bender2019,piantadosi2023, zhong2024opportunities}.

All these challenges highlight the need for a systematic methodological framework to guide the application of LLMs in language sciences. In response, the present study proposes an integrated framework that organizes three principal technical approaches: prompt-based interaction with general-use models, fine-tuning of open-source models, and embedding-based quantitative analysis. A comprehensive, normative methodological framework we proposed systematically aligns distinct LLM application techniques with explicit linguistic research objectives. This framework is designed to structure decisions about method selection, implementation, and evaluation, while explicitly addressing both technical and conceptual considerations. Based on this method-selection framework, we propose a systematic configuration framework to ensure seamless practical implementation by providing structured steps and clear parameter specifications within
a coherent research pipeline. Our two frameworks serve as an essential decision-theoretic guide, moving LLM-based inquiry from technical art to standardized scientific methodology. We not only delineate the two structured frameworks but also validate their efficacy and generalizability through a set of strict empirical studies, demonstrating their utility in generating, confirming, and mechanistically probing in language research.

The frameworks we proposed are distinguished by four essential features. First, they emphasize \textit{explicit methodology}, providing detailed specifications for each approach, including implementation details, parameter choices, and evaluation metrics, to support reproducibility. Second, they promote \textit{problem-method alignment}, guiding researchers to match research questions with appropriate technical approaches. Third, they incorporate a \textit{critical perspective}, highlighting the limitations of each method and identifying contexts where LLM-based approaches may be insufficient. Finally, they offer \textit{empirical demonstration} through concrete case studies with full methodological efficiency and transparency.

Although the proposed frameworks do not claim to resolve all debates about language research, they offer practical guidance for researchers seeking to incorporate LLM-based methods rigorously. By standardizing methodological reporting and fostering critical evaluation, they aim to facilitate cumulative knowledge building and enable meaningful comparison across studies. The present study proceeds as follows: Section 2 reviews current applications of LLMs in language sciences, Section 3 outlines the three methodological approaches, Section 4 offers detailed implementation guidance, Section 5 presents the integrative frameworks, Section 6 reports empirical validation of the proposed frameworks, and Section 7 discusses the implications and limitations.

\section{Current Applications of LLMs in Language Sciences}

LLMs have been applied across language sciences, spanning theoretical inquiry, applied research, and cognitive investigations \cite{katzir2023, koeva2024}. This section reviews current applications organized into four major domains, highlighting both achievements and methodological limitations. The recurring challenges across these domains highlight the need for a unified framework to ensure methodological coherence and theoretical interpretability.

\subsection{Linguistic structure and theory}

LLMs have been employed to investigate fundamental questions about language structure across multiple levels. For example, in syntax, language models like \texttt{BERT} have been probed for their representations of dependency relations, phrase structure, and agreement patterns \cite{htut2019,goldberg2019}. In semantics, contextualized embeddings enable study of polysemy, compositionality, and meaning relations \cite{periti2024systematic}. Pragmatic phenomena including implicature and sarcasm have been explored through classification tasks \cite{barbieri2020, ma2025}. Phonological modeling has utilized acoustic models to study sound patterns and phonotactic constraints \cite{baevski2020}.

However, significant interpretive challenges remain. When language models show sensitivity to grammatical patterns, this may reflect either internalization of linguistic rules or mere surface distributional regularities. The relationship between model behavior and theoretical linguistic constructs requires careful examination rather than direct equation. These interpretive ambiguities highlight the need for a unified methodological framework that systematically links theoretical aims with appropriate model interrogation strategies, ensuring consistent interpretation of linguistic evidence derived from LLMs.

\subsection{Language variation and social contexts}

LLMs facilitate large-scale analysis of language variation across social, geographical, and temporal dimensions. Applications include dialectal analysis from social media, tracking of language change, study of code-switching patterns, and cross-linguistic typological comparison using multilingual models \cite{eisenstein2014,pires2019}.

In applied linguistics and language education, LLMs support automated essay assessment, error detection in learner language, analysis of interlanguage development, and generation of pedagogical materials \cite{cong2025demystifying, neshaei2024towards, zhu2024concurrent}. Such tools facilitate large-scale corpus analyses of second language learning and individual learner paths.

Despite this, critical concerns include data bias. Training corpora disproportionately represent certain languages, varieties, and speaker populations while marginalizing others \cite{bender2019}. Applications in sociolinguistics and education must address whose language is represented and how models may perpetuate linguistic hierarchies. A unified framework can help mitigate such biases by establishing standardized criteria for data selection, representativeness, and model evaluation, thereby improving comparability and fairness in studies of linguistic variation.

\subsection{Language processing and cognition}

LLMs have been proposed as computational models of human language processing \cite{demszky2023,  niu2024}. For instance, \textit{surprisal} (= negative logarithm of word probability) computed by LLMs estimates from models correlate with reading times, eye movements, and neural activity patterns measured via fMRI and EEG \cite{schrimpf2021, hale2001, luo2025}. Research explores whether models can simulate acquisition trajectories, processing in clinical populations, and bilingual language use \cite{ren2024}.

In neurolinguistics, LLM representations have been aligned with brain imaging data to investigate neural encoding of linguistic information. These approaches aim to understand both computational mechanisms in models and biological mechanisms in brains.

However, fundamental differences exist between LLMs and human cognition. Models lack grounded perceptual experience, embodiment, social interaction, and the incremental processing characteristic of human comprehension. They process vastly more text than humans encounter. These differences constrain interpretation of models as direct cognitive analogs. A unified framework is thus needed to delimit the appropriate scope of model–human comparisons, ensuring that cognitive interpretations of LLM behavior are empirically grounded and methodologically consistent across studies.

\subsection{Corpus analysis and computational methods}

LLMs enable analysis of large-scale corpora through flexible pattern recognition, collocation detection, and automated annotation. They process unstructured text more readily than traditional concordance tools and facilitate diachronic analysis of semantic change \cite{hamilton2016diachronic}.

For low-resource and endangered languages, LLMs offer tools for documentation, transcription assistance, and pattern discovery in small corpora \cite{ginn2024can}. Multilingual models enable translation studies, contrastive analysis, and investigation of cross-linguistic correspondences.

Applications in discourse analysis include study of coherence relations, genre conventions, rhetorical structure, and argumentation patterns. However, the extent to which models capture discourse meaning versus surface correlations requires critical evaluation \cite{dentella2024}. Integrating these computational applications within a unified framework would provide standardized procedures for preprocessing, embedding extraction, and evaluation, ensuring that corpus-based findings derived from LLMs are reproducible and theoretically interpretable.

Taken together, these domains illustrate the transformative potentials of LLMs in language sciences, but also reveal a shared need for structured methodological coordination. The following section introduces a comprehensive framework that addresses these gaps through systematic method selection, implementation, and evaluation principles. 





\section{Overview of the Three Approaches}

Our proposed frameworks directly address the methodological fragmentation and other issues identified in the Introduction by providing systematic guidance across two dimensions: the method-selection framework (Sections 3 \& 4) and the constructed configurations framework (Section 5). 

LLMs can be applied in language sciences through three complementary strategies, each with distinct strengths and tradeoffs. First, \textit{prompt-based interaction with general-use LLMs} allows researchers to generate and analyze outputs from models like \texttt{GPT}-4 without accessing internal parameters. Second, \textit{fine-tuning open-source models} provides full control over architecture and weights, enabling task-specific adaptation and reproducible evaluation. Third, \textit{embedding-based quantitative analysis} extracts vector representations to study semantic relationships, detect large-scale patterns, or link linguistic features to behavioral and neural data. Choice of approach depends on research questions, technical resources, and desired balance between accessibility, reproducibility, and analytical depth (see Table~\ref{tab:comparison}). The three approaches form a method-selection framework of applying LLMs in language sciences.

\subsection{Approach 1: Prompt-based interaction with general-use models}

This approach queries closed-source LLMs such as \texttt{GPT}, \texttt{Claude}, or \texttt{Gemini} using natural language prompts or API calls \cite{kukreja2024literature, marvin2023prompt}, allowing researchers to elicit specific outputs and analyze them qualitatively or computationally. Since users cannot access model architectures, training data, or internal parameters, interaction is limited to text-based inputs and outputs, requiring only internet access and API credentials. Model behavior may change over time due to provider updates, which can affect the consistency of results.

Prompting ranges from single-turn queries for tasks such as grammaticality judgment or text generation to structured techniques like chain-of-thought reasoning, which uses multi-step instructions to improve output quality and interpretability \cite{wei2022}. API-based batch processing enables systematic application to larger datasets, and multi-agent frameworks coordinate multiple models or instances for complex analytical pipelines \cite{li2023}. Together, these methods provide a versatile toolkit for exploring a wide range of linguistic phenomena.

This approach is especially useful for generating exploratory hypotheses, rapidly prototyping analytical ideas, creating experimental stimuli, detecting preliminary patterns, and performing text classification or transformation when custom model training is impractical. Its low technical barrier and rapid iteration make advanced language capabilities accessible without training costs or local computational infrastructure.

However, outputs can be sensitive to prompt wording and stochastic sampling, and the internal mechanisms of closed-source models are opaque, making reproducibility and interpretability challenging. Commercial dependencies also raise concerns regarding long-term access, data privacy, and cost, while systematic evaluation is difficult without ground truth labels or clear metrics.

In short, employing prompt-based interactions with state-of-the-art general-use LLMs for the diagnostic assessment of their internal representations, analyzing linguistic phenomena (e.g., structural ambiguity resolution, pragmatic inference etc.), thereby facilitating the generation of testable linguistic hypotheses. Prompt-based interaction is most appropriate for exploratory research that prioritizes insight generation over reproducibility, proof-of-concept studies, pilot investigations, or situations lacking resources for model training. It is less suitable when reproducibility, understanding of model mechanisms, or very challenging analysis is required.

\subsection{Approach 2: Applying specialized open-source models}

This approach employs open-source specialized pre-trained models such as \texttt{BERT}, \texttt{RoBERTa}, or \texttt{T5}, which can be applied directly or fine-tuned on annotated datasets for classification, labeling, or generation tasks \cite{ding2023parameter}. Full model architectures and pre-trained weights are publicly available, allowing inspection and modification. Fine-tuning requires labeled data and computational resources, typically GPUs, and all hyperparameters can be controlled and documented to ensure reproducibility.

Direct application uses pre-trained models when existing capabilities suffice, whereas supervised fine-tuning adapts models to specific tasks such as sentiment analysis, grammaticality judgment, or error detection \cite{kastrati2025unlocking}. Parameter-efficient techniques like LoRA reduce computational costs \cite{hu2022lora}, and multi-task learning can improve generalization. These options provide flexibility for tailoring models to diverse research needs.

Applying fine-tuning on targeted, carefully-curated datasets to conduct confirmatory and theory-driven investigations. This approach allows for the empirical validation or refutation of specific linguistic theories for fine-grained analysis (e.g., constraints on word order, zero anaphora) by systematically manipulating training data and evaluating model performance against theoretical predictions.

Applications include supervised classification, sequence labeling, controlled text generation, and systematic corpus annotation. This approach enables robust quantitative evaluation, precise incorporation of domain knowledge, and model sharing within the research community. However, limitations include the need for technical expertise, labeled data, and computational resources, as well as potential overfitting. It is best suited for confirmatory studies, projects with annotated data, and research requiring reproducibility and strong empirical support.

\subsection{Approach 3: Embedding-based quantitative analysis}

The third application paradigm within our framework involves contextualized embedding. This method is not merely for quantitative data analysis, but serves as the dedicated pathway for ``probing the model's internal mechanisms'' and, critically, connecting them back to established linguistic theory.

Vector and contextualized embeddings provide the computational foundation for modern language models, representing words or tokens as high-dimensional numerical vectors. Contextualized embeddings, which vary depending on the surrounding context, encode information across multiple linguistic levels, including morphology, syntax, and semantics, while also partially reflecting phonological patterns and pragmatic cues \cite{camacho2018word}. By capturing these layers of linguistic information, embeddings enable models to perform complex language computations \cite{chang2019does}. Systematic exploration of embeddings through clustering, probing, and similarity analysis has revealed unexpected patterns and regularities, uncovering linguistic structures and tendencies that were difficult to detect with traditional methods. In this way, embeddings serve both as a technical mechanism for model computation and as a resource for empirical investigation, bridging theoretical linguistics and large-scale data-driven analysis \cite{garcia2021embeddings}.

By extracting the high-dimensional vector representations of linguistic units (words, phrases, or sentences) from various layers of an LLM, we transform the model from a black-box tool into a verifiable cognitive and linguistic probe. Our framework mandates that the analysis of these embeddings must go beyond mere data visualization (e.g., t-SNE) and must be explicitly designed to verify or challenge existing theories of language. For example, researchers can use embedding geometry (via geometric probing or representational similarity analysis) to test hypotheses regarding the hierarchical organization of syntactic structures, the compositionality of semantic representations, or the neurological plausibility of cognitive mechanisms hypothesized in psycholinguistics \cite{goldstein2024alignment}. 

In doing so, this approach elevates the use of LLMs from practical NLP (natural language processing) application to a novel form of computational theorizing, ensuring that the methodology yields results with deep and lasting theoretical significance for the language sciences. This approach scales efficiently to large datasets, facilitates hypothesis testing, and provides quantitative insight into phenomena traditionally studied qualitatively.

Extracting and analyzing contextualized embeddings via representational probing techniques to quantify the extent and location of encoded linguistic knowledge (e.g., syntactic tree structure, semantic role labeling) across the LLM's layers. This deep-dive mechanistic analysis utilizes formal mathematical tools to interpret model internal representations in light of established linguistic hierarchies. Whereas traditional linguistics relies on manual annotation, small datasets, and qualitative interpretation, it cannot easily capture large-scale semantic patterns, quantify subtle contextual shifts, or integrate multiple levels of linguistic information simultaneously. Contextualized embeddings, in contrast, enable data-driven analysis at unprecedented scale, tracking semantic change over time, comparing concepts across languages, discovering lexical relations automatically, analyzing sociolinguistic variation, and modeling aspects of human semantic memory. 


Challenges include the need for expertise in statistics and programming, difficulty interpreting embedding geometry, sensitivity to preprocessing and layer selection, and indirect mapping between embedding space and linguistic constructs. It is most effective for exploratory studies, gradient phenomena, large-scale pattern discovery, and linking linguistic features to behavioral or neural measures, but less suitable for categorical distinctions, small sample sizes, or cases where embeddings cannot be reliably linked to theoretical constructs.

\subsection{Comparison and selection criteria}

Table~\ref{tab:comparison} summarizes the main differences among the three approaches. Method selection should be driven by research questions, available resources, and acceptable tradeoffs between accessibility, reproducibility, and analytical depth. No single approach is universally superior, and each serves different research purposes.

\begin{table}
	\centering
	\caption{Comparison of Three Methodological Approaches}
	\label{tab:comparison}
	{\scalebox{0.95}{
			\small
			\begin{tabular}{@{}p{3cm}p{3.5cm}p{3.5cm}p{3.5cm}@{}}
				\toprule
				\textbf{Dimension} & \textbf{Approach 1: Prompting} & \textbf{Approach 2: Specialized Models} & \textbf{Approach 3: Embeddings} \\
				\midrule
				Technical barrier & Low & Medium to High & Medium to High \\
				Computational needs & Minimal (API only) & High (GPU required if fine-tuning) & Medium (CPU often sufficient) \\
				Reproducibility & Medium & High & Medium to High \\
				Transparency & Medium & High & Medium \\
				Development speed & Fast & Slow & Medium \\
				Data requirements & Minimal & Substantial labeled data(if fine-tuned) & Unlabeled text \\
				Primary output type & Text, qualitative & Classification metrics & Numerical measurements \\
				Best suited for & Exploration, generation & Confirmatory studies & Pattern discovery \\
				Required skills & Prompt design & ML engineering & Statistics, data science \\
				Cost structure & Per-query fees & Upfront infrastructure & Moderate compute time \\
				\midrule
				Research focus & \textbf{Rapid hypothesis generation, preliminary pattern detection} & \textbf{Externally verifiable outputs aligned with theoretical annotations} & \textbf{Internal representational structures} \\
				Methodological emphasis & Proof-of-concept investigations & Controlled, systematically verifiable outputs & Model as quantifiable cognitive system \\
				Theoretical validation & Medium & High (robust validation) & Deep insights \\
				Processing granularity & Coarse-grained & Fine-grained linguistic units & Variable-level analysis \\
				Primary use case & Early-stage exploration, content generation & Theory-driven confirmatory research & Understanding internal mechanisms and cognitive modeling \\
				\bottomrule
	\end{tabular}}}
\end{table}

Researchers should prioritize alignment between their specific research questions and the methodological affordances of each approach. The following section provides detailed implementation guidance to support informed application of these methods.

\section{Implementing the Three Approaches}

This section provides technical guidance for implementing each methodological approach, outlining design principles, procedural workflows, and interpretation frameworks necessary for robust application in language sciences research. The threeh approaches can form a practical and flexible framework that helps researchers select, combine, and apply methods effectively, ensuring both methodological soundness and adaptability to diverse research questions or tasks in language sciences.

\subsection{Approach 1: Prompt-based interaction with general-use models}

The deployment of closed-source LLMs through prompt-based interaction requires systematic design protocols, explicit documentation of all parameters, and critical evaluation of outputs against established linguistic knowledge \cite{giray2023prompt,marvin2023prompt}.

\textbf{Prompt design principles}: Effective prompts must balance specificity and flexibility. Three core components structure prompt design: (1) explicit task specification defining the linguistic phenomenon under investigation, (2) standardized input formatting ensuring consistency across instances, and (3) constrained output formatting facilitating systematic analysis.

Consider syntactic structure analysis as an illustrative case. The prompt should articulate the analytical task (``identify main and subordinate clauses''), provide the linguistic input in a consistent format, and specify the desired output structure (clause type, grammatical subject, main verb). This constrains the response space while permitting the model to apply its learned representations to the specific instance.

This strategy applies across linguistic subdomains. For phonological analysis, prompts may request pattern induction from underlying-to-surface form mappings. For pragmatic analysis, prompts may solicit identification of implicatures and the Gricean maxims they invoke. The critical requirement is explicit specification of the analytical framework guiding interpretation.

\textbf{Structured prompting for complex reasoning}: Complex linguistic analysis benefits from structured prompting techniques that decompose tasks into explicit reasoning steps. Chain-of-thought (CoT) prompting \cite{wei2022} requests intermediate reasoning, which can improve both output quality and interpretability \cite{zhou2022least}.

For ambiguity resolution, structured prompts may request several explicit steps. These include (1) enumeration of possible structural interpretations, (2) formal representation of each interpretation (such as phrase structure trees, dependency graphs, or logical forms), (3) evaluation of likelihood based on frequency, context, or processing constraints, and (4) theoretical explanation invoking relevant linguistic principles, including binding theory, prosodic phrasing, and discourse salience.

\begin{tcolorbox}[colback=gray!5!white,
	colframe=gray!60!blue,
	arc=3mm,
	boxrule=0.6pt,
	left=2mm,
	right=2mm,
	top=2mm,
	bottom=2mm,
	boxsep=1mm]
	
	\begin{quote}
		\textit{Prompt template:} Perform the following two tasks for the sentence below:
		
		\textbf{Subtask 1 – Clause Analysis:} Identify the main clause and all subordinate clauses. For each clause, specify the grammatical subject and main verb.
		
		\textbf{Subtask 2 – Syntactic Tree:} Using the \texttt{tikz-qtree} LaTeX package (or any LaTeX-compatible tree package), draw a syntactic tree of the sentence following a standard syntactic theory (e.g., X-bar theory, Minimalist Grammar, or Dependency Grammar). Ensure that all clauses and major phrase types are represented clearly.
		
		Sentence: The proposal that the committee member who the director had recommended submitted before the deadline was approved, although some reviewers who had initially opposed it changed their opinions.
		
		\textbf{Output Format:}
		\begin{itemize}[nosep]
			\item Main clause: [clause] (Subject: [X], Verb: [Y])
			\item Subordinate clause 1: [clause] (Subject: [X], Verb: [Y])
			\item Subordinate clause 2: [clause] (Subject: [X], Verb: [Y])
			\item Subordinate clause 3: [clause] (Subject: [X], Verb: [Y])
			\item Subordinate clause 4: [clause] (Subject: [X], Verb: [Y])
			
		\end{itemize}
		
	\end{quote}
	
\end{tcolorbox}

Consider the sentence, ``The student that the professor who the dean admired advised to revise the paper submitted it with hesitation''.

A structured prompt could elicit multiple types of analysis. First, it may enumerate ambiguities, such as the attachment of the prepositional phrase with hesitation'' (modifying submitted, ``revise'', or the entire complement clause), the interpretation of relative clauses (that the professor ... advised to revise the paper), and the control of pronouns across embedded clauses. Next, it can request formal representations, including phrase structure trees and dependency graphs, to capture each possible structural interpretation. Subsequently, the model can evaluate processing-based predictions, considering parsing preferences such as minimal attachment or late closure, and assess the likelihood of each attachment. Finally, the prompt can guide the model to provide a theoretical explanation, discussing binding theory for pronouns (it), prosodic phrasing effects on clause interpretation, and discourse salience considerations.

\textbf{Batch processing via Application Programming Interfaces (API)}: Systematic application to \textbf{a large amount of datasets} necessitates programmatic interaction through \texttt{API}s. The procedural workflow comprises several steps: (1) authentication and model version specification, (2) template construction with variable placeholders, (3) iterative instantiation across dataset instances, (4) parameter standardization (particularly temperature settings, which control output stochasticity), and (5) comprehensive metadata documentation.

The temperature parameter merits particular attention. Lower values (0.0 to 0.3) reduce sampling variance, increasing consistency across queries at potential cost to output diversity. However, even at temperature zero, most commercial APIs do not guarantee deterministic outputs due to implementation details. This fundamental indeterminacy constrains the reproducibility achievable through this approach.

Documentation requirements include: exact model identifier (e.g., ``gpt-4-0613'' rather than merely ``GPT-4''), all sampling parameters, query timestamps, API version, and any preprocessing applied to inputs or outputs. This metadata enables assessment of result stability and facilitates qualified replication attempts.

\textbf{Intelligent Agents (AI Agents)}: Closed-source LLMs are highly sensitive to prompt design and input order, with limited reproducibility of outputs. Building multi-agent systems (such as GPT-4, Gemini-2.5-flash, and Grok-3) can integrate the strengths of different models, enhancing analytical stability and functional coverage \cite{kapoor2024ai}. For example, GPT-4 can serve as a syntactic analyzer, \texttt{Gemini-2.5-flash} as a pragmatic interpreter, Grok as a phonological transcriber, and Claude as a content verifier, collaboratively processing the multidimensional features of the sentence ``The bird that the fox saw flew''.

\begin{tcolorbox}[colback=gray!5!white,
	colframe=gray!60!black,
	title={Sample Multi-Agent Output},
	fonttitle=\bfseries,
	arc=3mm,
	boxrule=0.6pt,
	left=2mm,
	right=2mm,
	top=2mm,
	bottom=2mm,
	boxsep=1mm]
	
	\paragraph{Agent A (GPT-4, syntactic analyzer):}  
	\textbf{Output:} The main clause is ``The bird flew,'' and the subordinate clause ``that the fox saw'' modifies the subject ``bird.''
	
	\paragraph{Agent B (Gemini-2.5-flash, pragmatic interpreter):}  
	\textbf{Output:} The sentence expresses a causal chain: the ``fox'' acts as perceiver, while the ``bird'' is the protagonist.
	
	\paragraph{Agent C (Grok-3, phonological analyzer):}  
\textbf{Output:} \textipa{/D@ "b3:d D@t D@ "fQks sO: flu:/}; primary stress falls on ``bird'' and ``fox.''
	
	\paragraph{Agent D (Claude-haiku-3.5, verification agent):}  
	Agent D integrates and cross-checks outputs to ensure consistency across syntactic, pragmatic, and phonological levels.
	
\end{tcolorbox}

Moreover, by employing frameworks such as \texttt{AutoGen}, and \texttt{LangChain} \cite{topsakal2023,guo2024}, researchers can coordinate role allocation, API invocation, and interaction logic, thereby achieving integrated analyses across syntactic, pragmatic, and phonological dimensions. However, implementing these frameworks requires strategies that extend beyond simple Python programming.

Immediate prompts, APIs, or AI agents all require engagement with prompt inputs, so we define this method as prompt-based interaction with general-use LLMs. This approach provides robust data-driven support for cross-level language modeling. A sample of the system output is shown below (code available at \url{https://osf.io/e6hjq}). General-use LLMs enable efficient language research via prompt design and APIs. Structured prompts support analysis of syntax, phonology, and pragmatics, while APIs allow large-scale corpus studies for dialogue, dialect, and cross-cultural analysis. Multi-agent systems enhance granularity and stability, linking linguistic structures to tasks without accessing internal model mechanisms. These prompt-based methods also apply to mainstream open-source general-use LLMs like \texttt{DeepSeek} and \texttt{Qwen}.

\begin{table}
	\centering
	\caption{Prompt-Based Interaction: Case Study (Exploratory Analysis)}
	\label{tab:prompt_case}
	{\scalebox{0.9}{
			\small
			\begin{tabular}{@{}p{3.5cm}p{8cm}p{3cm}@{}}
				\toprule
				\textbf{Component} & \textbf{Detail of Case Study} & \textbf{Reporting Standards} \\
				\midrule
				Research Question & What are the implicit semantic frames associated with the Chinese verbs \textit{shīluò} (\begin{CJK}{UTF8}{gbsn}失落\end{CJK}, ``lost/disappointed'') and \textit{shībài} (\begin{CJK}{UTF8}{gbsn}失败\end{CJK}, ``failed'') in contemporary internet discourse, and how do they differ? & \\
				\midrule
				Prompt Strategy & Zero-Shot Chain-of-Thought (CoT) Prompting. The model is instructed to first identify the most likely three-word context (agent, patient, or circumstance) for the target verb and then explain the resulting semantic frame before outputting the final frame name. & Prompt Engineering (CoT used), Temperature (Set to $T=0.7$) \\
				\midrule
				Data/Input & contextualized examples of each verb (e.g., ``He felt \textit{shīluò}.'') extracted from a specialized corpus of social media text. & Input Data (Source and size) \\
				\midrule
				Finding Example & GPT-4o consistently associated \textit{shīluò} with frames centered on internal, non-volitional states (e.g., `Empathy Gap,' `Self-Esteem Erosion'), while \textit{shībài} was linked to external, volitional, and measurable outcomes (e.g., `Competition Failure,' `Project Termination'). & \\
				\bottomrule
	\end{tabular}}}
\end{table}

Despite these advantages, three fundamental limitations constrain this approach. First, the opacity of model architectures and training data precludes a mechanistic understanding of their outputs. Second, while APIs enable processing of massive datasets, this does not guarantee output quality, particularly given the tendency of closed-source LLMs to \textbf{hallucinate}. As a result, they may be more suitable for tasks involving coarse-grained linguistic units rather than complex analyses. Third, sensitivity to prompt formulation introduces a degree of arbitrariness in the results. Moreover, although this approach can improve output coherence, it does not ensure theoretical validity. Model-generated analyses must be evaluated by experts against established linguistic frameworks. Consequently, the primary value of this methodology lies in the rapid generation of analytical hypotheses rather than in producing authoritative linguistic judgments.

Overall, this approach serves exploratory rather than confirmatory research functions. It enables rapid hypothesis generation, preliminary pattern detection, and proof-of-concept investigations.  However, it does not support strong claims about linguistic theory or cognitive processing without corroborating evidence from more rigorous methodologies. Table~\ref{tab:prompt_case} presents a case study illustrating prompt-based interaction for exploratory analysis. When research goals require high reproducibility, robust theoretical validation, and processing of fine-grained linguistic units, researchers may consider Approach 2, which involves fine-tuning or employing existing specialized language models. This alternative provides more controlled and systematically verifiable outputs.

\subsection{Approach 2: Applying specialized open-source models}

The current ecosystem of open-source models offers extensive resources for linguistic exploration. Platforms such as \texttt{HuggingFace} host more than two million publicly available models covering diverse architectures, domains, and languages. Researchers can often identify existing models that already perform annotation, classification, or generation tasks relevant to their studies. Employing such specialized models can significantly reduce computational costs and accelerate research. However, when existing specialized models do not align with the theoretical constructs or linguistic phenomena under investigation, such as specialized syntactic distinctions, pragmatic inferences, or lesser-studied languages, researchers may need to fine-tune models to achieve task-specific or theory-driven objectives.

Fine-tuning models for specific linguistic tasks enables reproducible, quantitatively evaluated research addressing well-defined questions \cite{lasri2022does,liu2025chain, alizadeh2023open,tan2024large}. This approach requires greater technical infrastructure but provides transparency and control absent from closed-source alternatives.

\textbf{Task formulation and dataset construction}: Rigorous application begins with precise operationalization of linguistic constructs as computational tasks. Common formulations include sequence classification (mapping texts to categorical labels), token classification (assigning labels to individual words or subwords), and sequence-to-sequence transformation (generating output texts from inputs).


Dataset partitioning follows standard machine learning practice: training (typically 60--80\% of data), validation (10--20\%), and test (10--20\%) sets. The training set optimizes model parameters. The validation set guides hyperparameter selection and early stopping. The test set, reserved until final evaluation, provides unbiased performance estimates. Critically, test set composition must not influence any modeling decisions to avoid overoptimistic performance estimates.


\textbf{Model selection and training procedures}: Pre-trained model selection should align with research objectives. BERT and its variants (RoBERTa, ELECTRA) excel at encoding tasks requiring bidirectional context, suitable for classification and sequence labeling. Encoder-decoder models (T5, BART) suit generation tasks. Multilingual models (mBERT, XLM-R) enable cross-linguistic analysis. Domain-specific models (BioBERT, SciBERT) may outperform general models on specialized corpora.

The training procedure involves several stages. First, tokenization converts text to model input format, requiring decisions about maximum sequence length (longer sequences increase computational cost but preserve context) and padding strategy (pad to maximum length versus batch-maximum length). Second, hyperparameter configuration specifies learning rate (typically $1 \times 10^{-5}$ to $5 \times 10^{-5}$ for fine-tuning), batch size (constrained by available memory), number of epochs (commonly 2--5), and optimization algorithm (AdamW is standard).

Third, training iterates through the dataset, updating model parameters to minimize loss on training examples while monitoring validation performance. Early stopping halts training when validation performance plateaus, preventing overfitting. Fourth, the best-performing checkpoint according to validation metrics is selected for final evaluation.

Reproducibility requires documentation of all decisions: base model and version, tokenization parameters, random seeds, hyperparameter values, hardware specifications (GPU model, memory), and training duration. Publishing code and trained models enables direct replication and facilitates cumulative science.





\textbf{Reporting standards for reproducibility}: Complete methodological reporting enables replication and facilitates meta-analysis across studies. Essential components include: model architecture specification (e.g., ``google-bert/bert-base-uncased'' from \texttt{HugginFace}), training data description (size, source, annotation procedure, inter-annotator agreement), complete hyperparameter specification, computational requirements (GPU hours, hardware specifications), evaluation metrics with confidence intervals, and error analysis results.
Table~\ref{tab:finetune_case} presents a case study illustrating fine-tuning for theory-driven investigation.

\begin{table}
	\centering
	\caption{Specialized Open-Source Language Models: Identification and Fine-Tuning}
	\label{tab:finetune_case}
	{\scalebox{0.85}{
			\small
			\begin{tabular}{@{}p{3.5cm}p{8cm}p{3cm}@{}}
				\toprule
				\textbf{Component} & \textbf{Details} & \textbf{Reporting Standards} \\
				\midrule
				Research Question & Can an open-source LLM, either directly or after fine-tuning, reliably distinguish between telic (goal-completed) and atelic (ongoing) event descriptions in a typologically diverse language (e.g., Swahili), and how does its performance compare to human annotators? & Clear definition of research objective \\
				\midrule
				Model Identification & Search existing specialized models on platforms such as \texttt{HuggingFace} for suitability to event-type classification. Select models already optimized for sequence classification or multilingual corpora. & Documentation of selection criteria, model name, version, and source \\
				\midrule
				Dataset Construction & Custom dataset of 3,000 Swahili sentences. Annotators: $N=3$ native speakers. Inter-Annotator Agreement (Cohen's $\kappa=0.85$). Dataset includes minimal pairs designed to capture theoretical contrasts. & Data size, annotation process, IAA, and theoretical coverage \\
				\midrule
				Fine-Tuning Procedure & LoRA (Low-Rank Adaptation) applied if model requires adaptation. Rank $r=8$, $\alpha=16$. Training: 5 epochs, learning rate $2 \times 10^{-5}$, weight decay 0.01. Early stopping based on validation performance. & Hyperparameters, adaptation method, training procedure documented \\
				\midrule
				Findings & Fine-tuned LLM achieved F1-score of 0.82. Performs well on typical sentences but shows errors in representing internal event structure for frequentative forms. Provides insight for theoretical interpretation and model refinement. & Critical evaluation, including limitations and areas of systematic error \\
				\bottomrule
	\end{tabular}}}
\end{table}

Nevertheless, interpreting the internal mechanisms of fine-tuned models remains challenging. For investigating how the model represents linguistic knowledge, researchers may turn to Approach 3 (Embedding-Based Quantitative Analysis).

\subsection{Approach 3: Embedding-based quantitative analysis}

Approach 2 aims to produce LLM outputs that align with theoretical annotations or predictions, focusing on the model’s behavior and its observable responses to tasks. In contrast, Approach 3 (Embeddings) investigates the internal representational structure of LLMs, treating the model as a quantifiable cognitive system and leveraging its internal variables to explain human behavior or linguistic phenomena. Together, these approaches provide a complementary framework: while Approach 2 emphasizes externally verifiable outputs, Approach 3 probes the internal mechanisms underlying those outputs, enabling deeper theoretical and cognitive insights.

Embedding extraction transforms linguistic data into numerical representations amenable to statistical analysis, enabling quantitative investigation of semantic, syntactic, and pragmatic phenomena \cite{tenney2019bert, jawahar2019does, reimers2019, giulianelli2020analysing, conneau2020}.

\textbf{Embedding extraction methods}: Contextualized embeddings represent words or sentences as high-dimensional vectors (typically 768 to 1024 dimensions) encoding distributional information from model training. Extraction requires several methodological decisions that affect downstream analysis.

Layer selection determines which model representations are extracted. Transformer models comprise multiple layers (12 or 24 commonly), each learning different abstractions. Early layers capture surface features and syntax, while later layers represent more abstract semantic information \cite{tenney2019}. Task requirements should guide selection: syntactic analysis may benefit from middle layers, while semantic similarity analysis often performs better with later layers.

Aggregation strategy converts token-level embeddings to sentence-level representations. Mean pooling averages embeddings across all tokens. Using the \texttt{CLS} token embedding (in BERT-style models) leverages the model's learned sentence representation. Max pooling captures the maximum value across tokens for each dimension. Weighted averaging applies importance weights based on attention mechanisms or term frequency. For autoregressive models like GPT, the last token embedding accumulates information from all previous tokens. Multi-layer aggregation combines embeddings from different layers through averaging or concatenation. These methods produce different representations; choice should be justified and documented.

The extraction procedure must be specified precisely: model name and version, specific layer(s) used, aggregation method, any normalization applied (unit normalization is common for similarity analysis), and contextual scope (isolated sentences versus sentences in document context).

\textbf{Similarity and distance analysis}: Quantifying relationships between embeddings typically employs distance or similarity metrics. Cosine similarity, ranging from $-1$ to $1$, measures angular similarity independent of vector magnitude. Euclidean distance captures absolute differences but is sensitive to vector norms. The choice depends on whether relative direction (cosine) or absolute position (Euclidean) better operationalizes the linguistic relationship of interest.

Similarity analysis proceeds through several steps: embedding extraction for all items of interest, computation of pairwise similarities, and interpretation relative to theoretical predictions or empirical baselines. Crucially, similarity values should be interpreted comparatively rather than absolutely. There is no universal threshold for high or low similarity; interpretation depends on the distribution of similarities within the dataset and theoretical expectations.


\textbf{Clustering and dimensionality reduction}: Unsupervised clustering algorithms partition embeddings into groups based on similarity, potentially revealing linguistic categories or patterns. Common algorithms include k-means clustering (requiring pre-specification of cluster number), hierarchical clustering (producing dendrograms showing nested groupings), and density-based methods like DBSCAN (identifying arbitrarily shaped clusters).

Cluster quality metrics assess whether discovered groupings are meaningful. Silhouette scores quantify how well instances fit their assigned clusters relative to other clusters. Davies-Bouldin index measures cluster separation and compactness. These metrics help select appropriate numbers of clusters and evaluate whether clustering reveals genuine structure versus artifacts.

Dimensionality reduction techniques project high-dimensional embeddings to two or three dimensions for visualization. t-SNE (t-distributed stochastic neighbor embedding, \cite{maaten2008}) preserves local neighborhood structure, emphasizing fine-grained distinctions. UMAP (Uniform Manifold Approximation and Projection) balances local and global structure. PCA (Principal Components Analysis) finds linear projections maximizing variance.


\textbf{Statistical modeling and hypothesis testing}: Embedding-based metrics can test specific hypotheses about linguistic phenomena through inferential statistics. The general framework involves: (1) extracting embeddings for relevant instances, (2) computing theoretically motivated metrics (similarity, clustering coefficients, dimensional projections), (3) applying appropriate statistical tests, and (4) interpreting results in linguistic terms.


Regression models can assess whether embedding-based metrics predict behavioral or neural measures while controlling for confounds. For instance, correlating embedding-based surprisal (derived from language model probabilities) with reading times while controlling for word length and frequency tests whether distributional information predicts processing difficulty.


\textbf{Downstream task applications}: Embedding-based analysis supports three primary analytical approaches: regression analysis for predicting linguistic outcomes, clustering for discovering categories, and dimensionality reduction for exploring structure.

Regression applications use embedding features to predict linguistic variables. Research might extract sentence embeddings to predict syntactic complexity scores, readability ratings, or processing times. The high dimensionality of embeddings requires careful treatment through principal components reduction or regularized regression methods.

Clustering applications partition linguistic items to discover or validate categories. Researchers might cluster word embeddings to identify semantic fields, or cluster sentence embeddings to categorize discourse functions. Validation is critical: compare discovered clusters to existing linguistic categories, assess cluster coherence through expert evaluation, or test cluster stability through cross-validation.


\textbf{Linking embeddings to linguistic constructs}: A fundamental interpretive challenge concerns the relationship between embedding space properties and theoretical linguistic constructs. Embeddings encode distributional co-occurrence patterns from training corpora. That two words have similar embeddings indicates they appear in similar contexts, not necessarily that they share semantic features, belong to the same syntactic category, or instantiate the same theoretical construct.

Interpretation therefore requires careful argumentation. Researchers must articulate why distributional similarity should correlate with the linguistic property of interest, acknowledge alternative explanations for observed patterns, and triangulate embedding-based findings with other evidence types (behavioral experiments, corpus analysis, theoretical predictions).

For example, high embedding similarity between metaphorical and literal uses of a word might reflect genuine semantic flexibility, or merely that both uses appear in diverse contexts. Disambiguating these interpretations requires additional analysis: examining which contexts drive similarity, comparing with explicitly polysemous words, or correlating with human similarity judgments, particularly useful in language variation, language changes, sociolinguistics, and language typology \cite{periti2024systematic, lucy2021characterizing, aljanaideh2020contextualized, grieve2025sociolinguistic}.



\begin{table}
	\centering
	\caption{Embedding-Based Explorations: Quantitative and Computational Analysis}
	\label{tab:embedding_case}
	{\scalebox{0.85}{
			\small
			\begin{tabular}{@{}p{3.5cm}p{8cm}p{3cm}@{}}
				\toprule
				\textbf{Component} & \textbf{Detail of Case Study} & \textbf{Reporting Standards} \\
				\midrule
				Research Question & Do contextualized representations of grammatical agreement markers exhibit consistent structural relationships across different layers, reflective of their hierarchical syntactic nature? & Theoretical Motivation \\
				\midrule
				LLM Used & BERT-base-uncased (12 layers, 768 dimensions) & Model Specification \\
				\midrule
				Data/Input & Controlled corpus of 1,000 minimal pairs (e.g., ``The key is on the table'' vs. ``The keys are on the table''), focusing on verb embeddings (is/are) & Input Data \\
				\midrule
				Extraction Method & Token-level embeddings extracted for target verbs from layers 6, 9, and 12 & Extraction Protocol \\
				\midrule
				Aggregation & No aggregation (token-level analysis); embeddings compared within-sentence across minimal pairs & Aggregation Strategy \\
				\midrule
				Analysis Method & Cosine similarity computed between agreement marker embeddings within each minimal pair; similarities compared across layers using repeated-measures ANOVA; PCA performed on embeddings from each layer & Statistical Method \\
				\midrule
				Results & e.g., Layer 9 showed highest within-pair similarity ($M = 0.89$, $SD = 0.08$) compared to Layer 6 ($M = 0.76$, $SD = 0.12$) and Layer 12 ($M = 0.71$, $SD = 0.15$); $F(2, 1998) = 247.3$, $p < 0.001$, $\eta^2 = 0.20$ & Quantitative Findings \\
				\midrule
				Interpretation & Mid-level layers are optimized for encoding local syntactic dependencies necessary for agreement, while earlier layers capture surface forms and later layers encode more abstract sentence meaning, supporting hierarchical processing accounts & Linguistic Interpretation \\
				\bottomrule
	\end{tabular}}}
\end{table}

\subsection{The method-selection framework: Bridging objectives and LLM paradigms}

The preceding sections outlined the implementation of the three methodological approaches. Based on these details and the model selection recommendations, a method-selection framework  naturally emerges. This practical and flexible framework organizes the three approaches (i.e., prompting, specialized or fine-tuning, and embeddings), according to their respective research functions and practical constraints (see Table~\ref{tab:comparison}). It provides researchers with structured guidance for selecting models and methods suited to specific tasks and research questions. In this sense, the framework serves as a decision roadmap that offers clear and adaptable criteria for choosing appropriate LLM strategies and implementation techniques in language science research.

To counter the prevailing methodological disarray, the core of this study is the method-selection framework, designed to guide researchers toward responsible and efficient decision-making. The value of this framework lies in its ``dual matching mechanism''.

\begin{enumerate}
	\item[] \textbf{Objective-driven matching:} The framework systematically categorizes linguistic research objectives into three distinct types: (1) \textit{exploratory analysis and hypothesis generation}; (2) \textit{confirmatory, theory-driven experimentation}; and (3) \textit{quantitative analysis and mechanism probing}. Crucially, the framework then provides a \emph{unique match} between each objective type and the most suitable LLM application paradigm (prompting, fine-tuning, or embedding). This structured approach ensures that the chosen method is inherently aligned with the scientific goal.
	
	\item[] \textbf{Resource and trade-off matching:} The framework clearly articulates the technical implementation details, inherent trade-offs (e.g., black-box vs. white-box access, data volume requirements, computational cost), and scope of applicability for each LLM method. This explicit resource-driven guidance ensures that researchers can efficiently select and deploy the correct methodology based on their available resources (e.g., data quality, computational power), thereby fundamentally preventing the inefficiency and errors caused by an objective-method mismatch.
\end{enumerate}

This method-selection framework is formalized as a \textbf{structured decision-pathway}, guiding researchers to the optimal LLM approach based on a multi-dimensional assessment of research objectives, available resources (data quantity, computational budget), and the required level of interpretability (e.g., choosing fine-tuning over prompting when high data efficiency is prioritized). 

Also the method-selection framework is intended as flexible guidance rather than a rigid prescription. Researchers are encouraged to adapt its recommendations to the demands of their specific projects. Method selection often involves balancing epistemological soundness, technical feasibility, and available resources. Thus, decisions should remain context-sensitive, informed by both theoretical aims and practical limitations. By treating the framework as a guiding structure rather than a fixed rulebook, scholars can make methodologically sound and context-appropriate choices.

At the core of this framework lies the principle that the research question and task complexity should drive the choice of methodology. Each method must align with the epistemological aim of the study, whether exploratory, confirmatory, or mechanistic. Selecting a method merely for convenience risks a mismatch between tool and inquiry, which can compromise both interpretability and validity. The primary goal is therefore to ensure that methodological robustness serves the research purpose.

While the method-selection framework provides a conceptual structure for choosing among LLM approaches, its focus remains on deciding which method best fits a given research question and constraint. In real-world applications, however, language science research often requires the integration of multiple methods across stages of inquiry. Researchers may need to combine exploratory prompting with targeted fine-tuning or embedding-based analysis in iterative cycles. This complexity calls for a more comprehensive framework that supports integration, workflow design, and methodological interaction beyond decision-making alone.

\section{Constructed Configurations Framework for LLM Applications}

Methodological guidance must ultimately translate into executable, reproducible practice. To this end, we propose the \textit{constructed configurations framework}, whose central value lies in providing a systematic and reproducible blueprint for the implementation of the ``multi-stage research pipelines'' common in language science. 

The method-selection framework serves primarily as a \textbf{conceptual guide} for choosing suitable approaches based on research goals, data, and resource constraints. However, not every study must adopt all three approaches, nor should they be seen as sequential steps. In many research settings, different methods are combined or iterated as the project develops.

To accommodate this need for flexibility, we propose a systematic framework for structuring LLM-based research in the language sciences. This framework extends the method-selection framework by emphasizing \textbf{method integration}, \textbf{workflow organization}, and \textbf{iterative refinement}. Rather than focusing solely on what approach to use, it provides an architecture for \textbf{how to connect and coordinate multiple approaches within a coherent research pipeline}. This systematic framework links exploratory, confirmatory, and representational analyses into an iterative process within a unified structure, thereby supporting the accumulation of reproducible and theoretically grounded findings.

This framework should be understood as a theoretically informed but adaptable blueprint rather than a fixed model. It aims to foster transparent, cumulative, and integrative research design by promoting interaction among methods and iterative refinement across stages of inquiry. In doing so, it provides a more comprehensive structure for applying large language models to empirical research in the language sciences.

This \textit{configuration framework} provides a catalogue of constructed configurations to guide multi-stage research pipelines.  Each configuration is defined by a formal notation:

\[
C = \text{Sequence}(M_{1} \rightarrow M_{2} \rightarrow \cdots \rightarrow M_{n}),
\]

where \( M_{i} \in \{\text{Prompt}, \text{Finetune}, \text{Embed}\} \), specifying a robust process flow. 
For example, the refinement configuration can be represented as:

\[
C_{\text{Refinement}} = \text{Prompt} \rightarrow \text{Finetune} \rightarrow \text{Embed}.
\]

Our configuration framework ensures seamless practical implementation by providing structured steps, clear parameter specifications, and explicit transition logic between methods. This systematic approach guarantees two key outcomes:

\begin{enumerate}
	\item[] \textbf{Practical implementation efficacy:} Researchers can use the framework's visual and textual blueprint to rapidly construct and execute complex LLM research pipelines, significantly lowering the technical barrier to entry.
	
	\item[] \textbf{Verifiable reproducibility:} Each constructed configuration represents an independently verifiable unit. This systematic standardization of the research process provides a robust guarantee for the ``experimental reproducibility'' that the language science community urgently requires, directly addressing the core challenge of the reproducibility crisis.
\end{enumerate}


\subsection{Framework architecture and design principles}

The systematic framework organizes the methodological approaches into a modular architecture guided by three core principles.  
First, each approach can be deployed independently or combined with others, depending on research objectives, theoretical orientation, and computational resources.  
Second, findings obtained through one approach can inform and refine subsequent analyses, supporting iterative development of hypotheses and models.  
Third, method selection and integration should be explicitly grounded in research questions and evidentiary standards within specific subfields of language sciences.

This architecture reflects established practice in computational research but contributes by making the logic of integration explicit and accessible to scholars less familiar with technical methodologies. The goal is not to prescribe a rigid pipeline but to clarify how modular components interconnect, enabling transparent, reproducible, and theoretically interpretable research designs.

In the schematic representation of the framework, the three approaches form interlinked modules, with arrows indicating possible directions of analytical progression and feedback. Nevertheless, actual research rarely follows such a linear sequence. Unexpected findings, practical limitations, or theoretical debates may necessitate deviations from the idealized flow. The framework should not be taken as a strict procedural rulebook.

\subsection{Two prototypical configurations}

\begin{figure}
	\centering
	\includegraphics[width=0.8\textwidth]{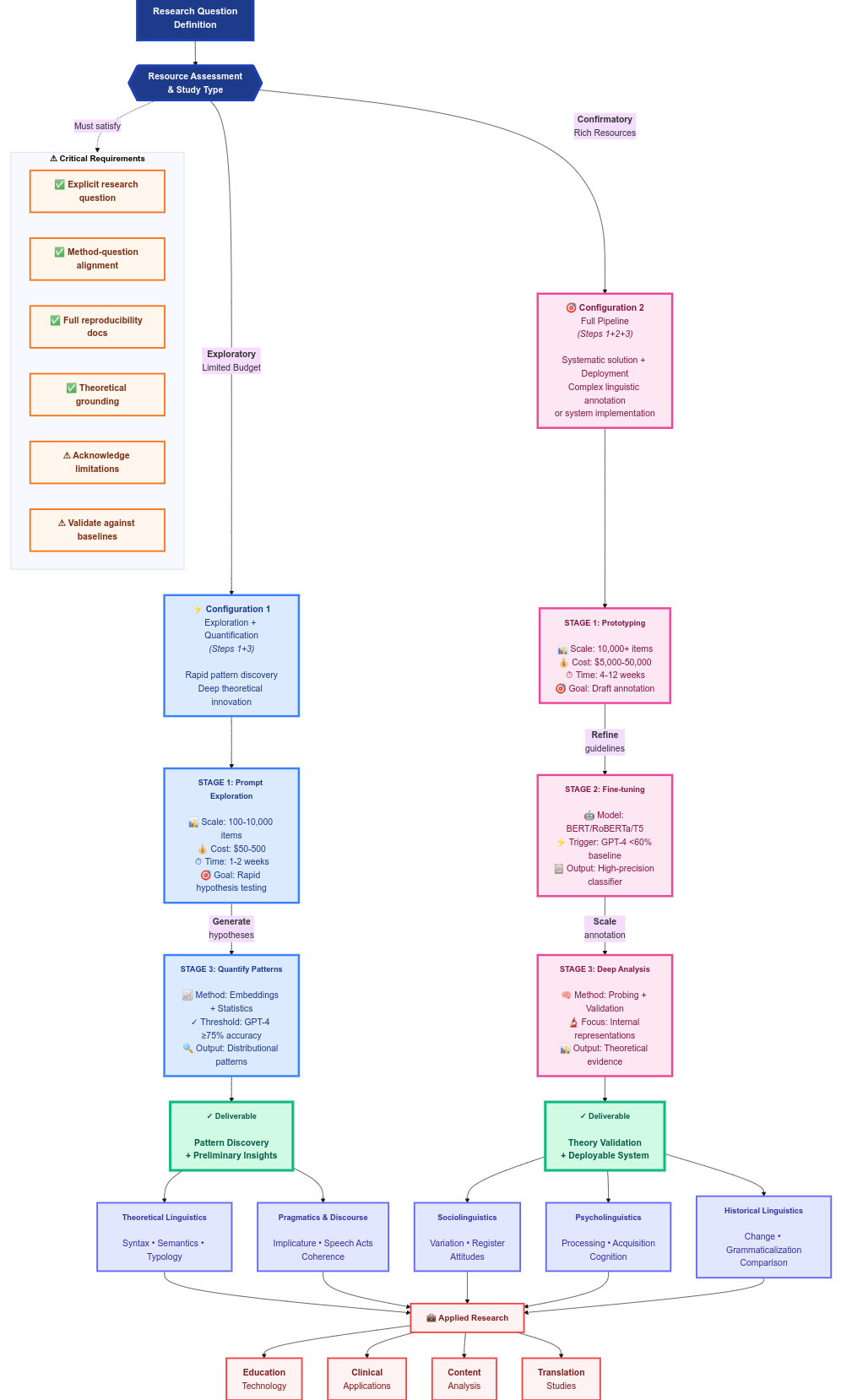}
	\caption{Systematic Framework for LLM-based Language Sciences.
		{\footnotesize{Note: This framework illustrates a dual-path approach that aligns research design with resource conditions and study purposes. The left path (Configuration 1) supports exploratory studies with limited resources, emphasizing rapid pattern discovery and theoretical innovation through prompt-based hypothesis testing and quantitative analysis. The right path (Configuration 2) represents a confirmatory route suited for resource-rich settings, integrating full-scale prototyping, fine-tuning, and deep model analysis to yield validated theories and deployable systems. Both configurations share a common foundation of critical research requirements and lead to diverse applications across theoretical and applied linguistic domains.}}}
	\label{fig:flowchart}
\end{figure}

To demonstrate how this modular system can be applied, we outline two prototypical configurations that combine the proposed approaches in complementary ways. These are simplified examples meant to illustrate typical integration patterns rather than prescribe fixed procedures. In practice, implementations are dynamic, involving iteration, reanalysis, and methodological adaptation. The general structure of these configurations is shown in Fig.~\ref{fig:flowchart}, offering concrete reference points for designing multi-stage or hybrid LLM studies in linguistics.

\subsubsection{Configuration 1: Exploration and quantification (Approaches/Stages 1 + 3)}

This configuration combines prompt-based exploration with embedding-based analysis, omitting the resource-intensive fine-tuning stage. It suits exploratory research where rapid iteration is prioritized over maximum reproducibility.

\medskip
\noindent\emph{Workflow.} Stage 1 uses general-use LLMs to generate hypotheses or preliminary classifications. Stage 3 applies embedding-based methods to quantify patterns suggested by Stage 1. Results from Stage 3 may validate, contradict, or complicate Stage 1 findings.

\medskip
\noindent\emph{Advantages.} Minimal infrastructure requirements, rapid exploration of novel phenomena, and efficient scaling to large corpora. 

\medskip
\noindent\emph{Limitations.} Reproducibility constraints from Stage 1, lack of custom model training that might better capture domain-specific patterns, and absence of systematic evaluation metrics from supervised learning.

\medskip
\noindent\emph{Appropriate use cases.} Preliminary investigations where creating annotated training data is impractical. \emph{Inappropriate use cases.} When reproducibility is essential, when strong empirical claims require robust support, or when the phenomenon demands precision that general-purpose models cannot provide.

\medskip
\noindent To sum up, \texttt{Configuration~1} is suitable for small-scale investigations that emphasize \emph{rapid pattern discovery} and \emph{deep theoretical innovation}. The linguistic granularity in this configuration is relatively coarse, focusing on features such as part-of-speech categories, basic grammatical structures, and broad emotional dimensions. It is particularly well suited for studying representative linguistic phenomena in areas such as pragmatics (e.g., implicature), syntax (e.g., dependency relations), and sociolinguistics (e.g., gender-related variation).

\vspace{0.5cm}
\textbf{Application example 1: Emoji effects on utterance interpretation}

\begin{quote}
	\small
	\emph{Methodological note:} This example illustrates the configuration but should not be taken as definitive evidence that the approach works well for this question. Alternative methods might yield different or contradictory findings.
\end{quote}

\noindent\fcolorbox{blue!60}{blue!5}{%
	\parbox{0.95\textwidth}{%
		\textsc{\color{blue!80}Research Question:} 
		\textit{How do emoji additions modulate pragmatic interpretation of utterances?} 
		\cite{herring2024emoji, liu2021improving}
	}%
}

\smallskip
\noindent\fcolorbox{green!60}{green!5}{\textsc{\color{green!70!black}Stage 1: Exploration}}

\smallskip
Construct 500 neutral utterances paired with diverse emojis. Query GPT-4 via API to generate interpretations of each combination, requesting assessments of speaker attitude and emotional valence.

\emph{Methodological issues:} Model outputs vary with prompt wording and temperature settings. Different queries of the same combination can yield inconsistent interpretations. The model's understanding of emojis may reflect training data patterns rather than pragmatic principles. We cannot determine whether outputs reflect genuine pragmatic reasoning or surface pattern matching.

\emph{Preliminary patterns:} 
Positive symbols (+) tend to amplify positive interpretations, 
whereas negative symbols (-) often introduce ironic or sarcastic readings. 
The consistency remains nice: approximately 75\% of outputs contradict this trend or yield ambiguous interpretations. 
Even when incorporating an additional LLM (e.g., Gemini) as a validation agent to improve precision, 
overall consistency remains below 85\%. This means that the classifier task could be done well by LLMs. However, we want to have further analysis quantitatively.

\smallskip
\noindent\fcolorbox{orange!60}{orange!5}{\textsc{\color{orange!80!black}Stage 3: Quantification}}

\smallskip
Extract BERT embeddings for each utterance-emoji combination. Compute cosine similarities to prototypically positive, negative, and neutral expressions. Conduct statistical tests comparing similarity distributions.

\emph{Results:} Utterances with positive emojis show higher similarity to positive expressions ($M = 0.84$, $SD = 0.09$) than baseline ($M = 0.61$, $SD = 0.12$; $t(498) = 12.4$, $p < 0.001$, $d = 1.1$). Negative emoji combinations cluster with negative/ironic expressions ($M = 0.71$ vs.\ baseline $M = 0.48$; $t(498) = 9.8$, $p < 0.001$, $d = 0.9$).

\smallskip
\noindent\textsc{Critical evaluation}

\emph{Interpretation challenges:} These findings demonstrate that emoji-modified utterances occupy different regions of BERT's embedding space. However, what this means for pragmatic interpretation is unclear. The correlation between embedding similarity and human pragmatic judgment requires validation through behavioral experiments. BERT was not trained to represent pragmatic meaning; observed patterns may reflect superficial co-occurrence rather than pragmatic principles.

\emph{Limitations:} No human judgment data validates model interpretations. No comparison with alternative methods (human annotation, behavioral experiments) demonstrates whether this approach yields insights unavailable through other means. The statistical effects, while significant, do not establish that the approach successfully addresses the research question in a theoretically meaningful way.

\subsubsection{Configuration 2: Comprehensive workflow (Approaches/Stages 1 + 2 + 3)}

This configuration integrates all three approaches into a full pipeline. It addresses complex questions requiring multiple evidence types but demands substantial resources and expertise.

The workflow begins with Stage 1 exploration, proceeds to Stage 2 supervised learning with annotated data, and concludes with Stage 3 pattern analysis across larger corpora. In practice, the process is rarely linear. Annotation difficulties may require returning to Stage 1 for revised operationalizations. Model failures may reveal that the phenomenon is not computationally tractable as formulated. Stage 3 may uncover unexpected patterns that contradict Stage 2 results.

This configuration provides multiple forms of evidence but does not guarantee valid conclusions. Each stage introduces potential errors: Stage 1 from model inconsistencies, Stage 2 from annotation decisions and modeling choices, Stage 3 from interpretation of distributional patterns. Errors can compound rather than canceling out.

To sum up, Configuration 2 is suitable for two typical cases in language sciences. The first involves \textit{complex linguistic phenomena} that require large-scale investigations with fine-grained linguistic granularity and intricate patterns, such as distinctions between prepositions and implicatures, sarcasm detection, and the analysis of zero anaphora \cite{sun2019}. The second concerns \textit{system implementation}, where the research goal is to build systems capable of automatically identifying or precisely detecting linguistic phenomena in language data, such as diagnosing language disorders in children or detecting toxic content in children's books. Table~\ref{tab:config2} illustrates this comprehensive configuration focused on maximizing data quality and aligning LLM capabilities with established theoretical frameworks.

\begin{table}
	\centering
	\caption{Prototypical Configuration 2: Theory Validation and Annotation}
	\label{tab:config2}
	{\scalebox{0.85}{
			\small
			\begin{tabular}{@{}p{2.5cm}p{3.5cm}p{5cm}p{4cm}@{}}
				\toprule
				\textbf{Stage} & \textbf{Approach Used} & \textbf{Goal} & \textbf{Output} \\
				\midrule
				Stage 1 (Rapid Prototyping) & Prompt-Based Interaction (Approach 1) & To rapidly generate a pilot dataset of linguistically ambiguous or complex examples based on the researcher's specific theoretical criteria (e.g., identifying subtle differences between near-synonyms). & A high-volume, initial Draft Annotation Set (low confidence). This set allows the researcher to refine annotation guidelines and identify edge cases. \\
				\midrule
				Stage 2 (Refinement and Validation) & Fine-Tuning (Approach 2) & To train a high-precision, reproducible classifier that can validate the human annotation process and potentially scale the annotation. & A High-Confidence, Fine-Tuned Model (e.g., $F1 > 0.90$) and a Gold-Standard Annotated Corpus (where model predictions match human annotations). The model serves as a ``reproducible annotator'' for future data. \\
				\midrule
				Stage 3 (Mechanism Insight) & Embedding-Based Analysis (Approach 3) & To use the gold-standard corpus to probe the model's internal representations, confirming whether the model's internal linguistic processing aligns with the human-derived linguistic theory. & Vector Visualization (e.g., t-SNE plot) showing that the embeddings for the theory-relevant features cluster distinctly, thereby providing evidence for the representational validity of the linguistic construct within the model. \\
				\bottomrule
		\end{tabular}}}
	\end{table}
	
	This iterative process is crucial for resource-constrained or theory-focused linguistics. It uses the LLM's speed for initial data generation (Stage 1), its reproducibility for validation and scaling (Stage 2), and its internal structure for deep theoretical insight (Stage 3).
	
	\vspace{0.5cm}
	
	\textbf{Application example 2: Topic tracing in Chinese discourse}
	
	\begin{quote}
		\small
		\emph{Methodological note:} This example demonstrates integration of all three stages but highlights the challenges that arise when combining multiple imperfect methods.
	\end{quote}
	
	\noindent
	\fcolorbox{blue!60}{blue!5}{%
		\parbox{0.95\textwidth}{%
			\textsc{\color{blue!80}Research Question:} 
			\textit{How can topic introduction, maintenance, and reintroduction be automatically identified in Chiense discourse?} 
			\cite{sun2019, wei2024llms, zhang2021edtc}
		}%
	}
	
	\smallskip
	\noindent\fcolorbox{green!60}{green!5}{\textsc{\color{green!70!black}Stage 1: Hypothesis generation}}
	
	\smallskip
	\noindent\fcolorbox{blue!60}{blue!5}{\textsc{\color{blue!80}Research Question:}} \quad \textit{How can topic chains with zero anaphora be automatically identified in Chinese discourse?} 
	
	\medskip
	
	Topic chains refer to sequences of clauses connected by zero anaphora (phonetically null pronouns) in Chinese discourse. For example:
	
\begin{CJK}{UTF8}{gbsn}
小明去了商店。 {Ø} 买了苹果。{Ø} 回家了。
\end{CJK}

\textit{Xiaoming went to the store. \underline{[He]} bought apples. \underline{[He]} went home.}

	\noindent The subject ``Xiaoming'' appears only once but is implicitly understood across all three clauses (marked as Ø). This phenomenon is quite challenging for computational and quantitative research because: (1) zero anaphora lacks overt linguistic markers, (2) topic boundaries are often ambiguous, and (3) resolution requires deep pragmatic understanding \cite{li1976subject}.
	
	However, with the aid of LLMs, we can potentially facilitate data-driven research on this topic. LLMs can: (1) automatically identify zero anaphora positions through contextual inference, (2) resolve coreference chains by tracking implicit referents, and (3) enable large-scale corpus annotation for quantitative analysis \cite{sun2019, wei2024llms, zhang2021edtc}. 
	
	First, sample the 200 texts with discourse chains. Use GPT-4 to identify topic chains, requesting classification of mentions as introduction, continuation, or reintroduction and description of linguistic devices.
	
	\emph{Challenges encountered:} Output consistency varied considerably. Across 50 segments analyzed three times with identical prompts, agreement between runs averaged only 30.2\% (Fleiss’ $\kappa = 0.24$), suggesting low reliability. Certain patterns appeared genre-specific, yet the models failed to consistently indicate which patterns were generalizable. Incorporating two additional LLMs (e.g., Gemini, Grok) into an ensemble or agent setup increased agreement to about 38.6\%. Despite this modest improvement, such annotations remain unsuitable for direct research use.
	
	These inconsistencies make it impractical to employ Stage~1 outputs as finalized annotations. Instead, we use them to identify candidate phenomena for Stage~2 annotation, treating Stage~1 primarily as a hypothesis-generation phase rather than as a preliminary labeling step.
	
	\smallskip
	\noindent\fcolorbox{purple!60}{purple!5}{\textsc{\color{purple!80}Stage 2: Model development}}
	
	\smallskip
	Create annotated training data with 1,000 texts. Expert annotation following guidelines refined from Stage 1 insights. Inter-annotator agreement (Fleiss' kappa $= 0.78$) indicates reasonable but not perfect reliability, with persistent disagreements on ambiguous reintroductions. Fine-tune \texttt{RoBERTa} for sequence labeling. Performance on held-out test data: $F1 = 0.81$ for continuation, $F1 = 0.74$ for introduction, $F1 = 0.69$ for reintroduction.
	
	\emph{Limitations:} Performance varies by genre (news $F1 = 0.83$; conversation $F1 = 0.72$). The model struggles with ambiguous cases that challenged human annotators. Error analysis reveals systematic failures on discourse segments with multiple competing topics, suggesting the model relies on local cues rather than global discourse representation.
	
	
	\smallskip
	\noindent\fcolorbox{orange!60}{orange!5}{\textsc{\color{orange!80!black}Stage 3: Large-scale analysis}}
	
	\smallskip
	Apply trained model to 5,000 texts. Extract embeddings for topic-referring expressions grouped by predicted role. Analyze similarity patterns.
	
	\emph{Findings:} Topic continuations show higher similarity to antecedents ($M = 0.87$) than reintroductions ($M = 0.72$; $t(48998) = 43.2$, $p < 0.001$, $d = 0.61$). Similarity decays during topic gaps and partially recovers at reintroduction. Genre differences emerge in topic management strategies.
	
	\emph{Complications:} These patterns hold for model predictions, but given Stage 2's 0.69--0.81 F1 scores, approximately 20--30\% of predictions are incorrect. Patterns observed in Stage 3 may partially reflect model biases rather than true discourse structure. Distinguishing genuine linguistic patterns from modeling artifacts requires additional validation that this study did not perform.
	
	\smallskip
	\noindent\textsc{Integration and overall assessment}
	
	Stage 1 identified relevant phenomena and generated hypotheses. Stage 2 developed a working system with quantified performance. Stage 3 revealed distributional patterns. However, multiple limitations accumulate: Stage 1 inconsistency, Stage 2 errors especially on complex cases, and Stage 3 interpretive challenges. The research provides useful preliminary findings but falls short of definitive conclusions about topic management in discourse.
	
	
	\subsection{Constructed configurations framework for method selection}

	We proposed this constructed configurations framework that integrates epistemological orientation, methodological design, and practical constraints into a coherent process (Fig.~\ref{fig:flowchart}). The goal is not to prescribe a single best practice, but to provide a transparent structure that helps researchers make principled and reproducible methodological choices.
	
	This systematic framework begins with the formulation of a clear research question, followed by an assessment of available resources and study type. This step determines whether the study is exploratory, typically constrained by limited data or budget, or confirmatory, supported by richer resources. A set of critical requirements, including an explicit research question, alignment between method and research goal, theoretical grounding, reproducibility documentation, and acknowledgment of limitations, must be satisfied before proceeding to implementation.
	
	Two operational configurations are defined within this framework. 
	Configuration~1 supports exploratory inquiry with limited resources, emphasizing prompt-based exploration and quantitative analysis to identify distributional patterns and generate hypotheses. 
	Configuration~2 applies to confirmatory studies with sufficient resources, involving a complete pipeline of prototyping, fine-tuning, and deep analysis aimed at theory validation or deployable system construction. 
	Although the framework suggests a linear progression between these configurations, research practice often involves iteration and adaptation in response to theoretical or practical challenges.
	
	The underlying principle is that methodological choice should be driven by the research question and aligned with the epistemological aim of the study. Exploratory investigations benefit from flexible, rapid methods, whereas confirmatory inquiries require thoroughness, annotation quality, and reproducibility. When constraints prevent full methodological soundness, researchers should explicitly frame their study as exploratory and limit claims accordingly. Transparency about trade-offs and limitations is central to the framework’s ethical foundation.
	
	Table~\ref{tab:selection} summarizes crucial considerations for selecting between the two operational configurations. These are intended as flexible guidelines rather than fixed prescriptions, as appropriate methodological choices depend on research objectives, available resources, and disciplinary conventions.
	
	\begin{table}
		\centering
		\caption{Considerations for Selecting Operational Configurations}
		\label{tab:selection}
		\small
		\begin{tabular}{@{}p{3cm}p{5.5cm}p{5.5cm}@{}}
			\toprule
			\textbf{Consideration} & \textbf{Configuration 1 (Stages 1+3)} & \textbf{Configuration 2 (Stages 1+2+3)} \\
			\midrule
			Research phase & Exploratory & Confirmatory \\
			Primary goal & Pattern discovery and hypothesis generation & Robust validation and system deployment \\
			Resources needed & Moderate & Substantial \\
			Data availability & Unannotated or public corpora & Annotated or annotatable data \\
			Timeline & Weeks to months & Months to years \\
			Reproducibility & Moderate, limited by prompt variability & Higher, supported by documented pipelines \\
			Technical expertise & Moderate & High \\
			Risk of failure & Moderate (limited scope) & High (complex dependencies) \\
			\midrule
			\textbf{Appropriate when} & Studying new or poorly understood phenomena; resources are limited; rapid iteration needed; annotated data unavailable & Empirical validation or theoretical testing required; reproducibility essential; sufficient data and computational capacity available \\
			\midrule
			\textbf{Inappropriate when} & Strong theoretical claims or high reproducibility standards required & Exploratory insights or flexible experimentation needed; annotation infeasible; research question not precisely defined \\
			\bottomrule
		\end{tabular}
	\end{table}

	\subsection{When not to use this framework}
	
	This framework has several inherent limitations that constrain its applicability. 
	For instance, there is a risk that researchers may apply the framework mechanically rather than critically adapting methods to specific questions, and no empirical evidence yet demonstrates that it improves research quality relative to alternative approaches. Its focus on computational techniques may also could potentially overshadow traditional linguistic analysis, experimental methods, or fieldwork, which are often more suitable for certain research questions.
	
	Moreover, LLM-based approaches are therefore best avoided when the phenomenon of interest involves situated meaning, embodied experience, or social context, when sample sizes are small and qualitative analysis is more informative. They may also be inappropriate if computational resources would be better devoted to human annotation or experimental studies, or if the norms and evidentiary standards of the research community do not recognize computational evidence as sufficient. Overall, the framework should be understood as one way to organize methodological options rather than a comprehensive approach to language sciences, with its utility depending on the nature of the phenomenon, the data, and the research context.

	\section{Empirical Validation of the Framework}
	
	\subsection{Validation strategy and measurement approach}
	
	Having presented our two systematic frameworks for LLM-based language research (Sections 3--6), we now evaluate their practical utility through retrospective analysis of published studies. The validation addresses a critical question: \textit{Does applying our framework's specifications demonstrably improve methodological efficiency and transparency?}
	
	To answer this question, we operationalize ``methodological transparency'' through a quantitative assessment instrument that directly maps to our framework's components. Each of the framework's five specification domains (Sections 4.1--4.5): \textit{model specification, parameter documentation, data description, reproducibility elements, and limitation acknowledgment}, becomes a measurable dimension in our evaluation. This mapping ensures the 
	validation directly tests whether following framework guidelines improves documentation quality in the areas we identified as critical.
	
	More importantly, we conducted a three-stage validation of methodological efficiency to demonstrate the robustness of the proposed framework. The validation method includes: (1) retrospective analysis, which examines existing studies to identify documentation gaps; (2) prospective analysis, which applies the framework to replicate a previous experiment and assess its practical usability; and (3) expert validation, which invites domain experts to evaluate the framework’s contribution to transparency and interpretability. Together, these stages provide converging evidence that the framework strengthens methodological soundness, accountability, and reproducibility in LLM-based language research.

	
	
	
	
	
	
	
	
	
	\subsection{Validation method and results}
	
	To evaluate the practical value and generalizability of the proposed framework for applying LLMs in language sciences, we conducted a three-stage validation combining retrospective analysis, prospective replication, and expert evaluation. This design follows a before–after logic: assessing existing studies in their original form, then reassessing how their methodological transparency and soundness would improve under framework guidance.
	
	\textbf{Method 1: Retrospective documentation analysis.} We first assessed a representative set of published studies using their original documentation, scoring each along the five core dimensions of Methodological Transparency Scale (MTS): model selection rationale, task–model alignment, transparency of data handling, reproducibility of results, and theoretical interpretability. These baseline scores reflected the status of current practice without systematic framework guidance. We then mapped each study to the appropriate methodological pathway (e.g., prompting, fine-tuning, or embedding-based) defined in Sections~4–6 and re-evaluated what modifications would occur under framework compliance. The difference between original and framework-aligned scores thus quantified the extent to which the framework addresses real documentation gaps.
	
	\textbf{Method 2: Prospective framework application.} To test usability beyond retrospective scoring, we implemented a forward-looking experiment by re-running an existing LLM-based study on sentence-level semantic similarity and reading speed prediction. The replication adhered strictly to the framework’s stepwise specifications, including explicit documentation of decision points, comparison of baseline models, justification for embedding choice, and transparency of evaluation metrics. The framework-guided version achieved clearer methodological traceability and improved reproducibility, confirming that the framework is not only diagnostic but also practically operational.
	
	\textbf{Method 3: Expert validation survey.} Finally, to evaluate the framework’s perceived value among specialists, we conducted an expert survey with 38 researchers in NLP, computational linguistics, and cognitive science. Each participant reviewed two versions of the same study: one written in the conventional descriptive style and one structured according to our framework. They rated both versions across six dimensions (i.e., model selection justification, clarity, replicability, transparency, articulation of limitations, and overall quality) on five-point Likert scales. Results showed significant improvement in all dimensions, with $95\%$ of respondents preferring the framework-guided version. Qualitative comments emphasized its strengths in explicating rationale, ensuring decision traceability, and promoting consistent methodological reasoning.
	
	The three validation methods provide complementary evidence for the framework’s robustness. The following section describes the implementation of each method used to test the validity of our approaches and presents the corresponding results. 

	\subsubsection{Retrospective analysis: Preventing attribution errors}
	
	
	A recurrent pitfall in computational linguistics is the tendency to interpret strong task performance as evidence of genuine linguistic understanding. Consider a study (Research E) that fine-tunes an open-source LLM on a discourse coherence classification task and reports an F1 score of $0.92$. The authors infer from this metric that the model has acquired a specific \textit{discourse reasoning mechanism (DRM)} rooted in linguistic theory. However, such claims often rely on behavioral performance alone and fail to examine the underlying representations or causal mechanisms responsible for the observed behavior.
	
	This issue parallels the ``right for the wrong reasons'' problem identified by \cite{mccoy2019right}. In their analysis of natural language inference (NLI), models such as BERT seemed to exhibit syntactic generalization, but probing and adversarial tests revealed that their success stemmed from shallow lexical heuristics and dataset biases rather than genuine syntactic competence. Extending this insight, \cite{tenney2019bert} showed that BERT’s representations follow a hierarchical structure: lower layers encode morphology and syntax, while higher layers capture semantics. This finding shows that high task performance does not necessarily reflect the acquisition of theoretically relevant mechanisms. More recently, \cite{uluslu2025investigating} found similar misalignments in native language identification tasks, where LLMs achieved strong results by relying on cultural or self-referential cues instead of linguistic structure. Collectively, these studies highlight the risk of equating performance with understanding and motivate the need for explicit mechanism-level validation to ensure interpretability and theoretical soundness.

	The proposed framework directly prevents such attribution errors by enforcing a mechanism-validation stage after model fine-tuning. Specifically, it mandates a transition from \textbf{Stage 2} (specialized open-source models) to \textbf{Stage 3} (embedding representation analysis), where model-internal evidence is systematically examined. The framework introduces a standardized \textit{representation-probing protocol}, for instance: \textit{Model: fine-tuned LLM from Stage 2; Layer: 9; Token: [CLS]; Metric: cosine similarity between categories}. This protocol enables testing alternative hypotheses such as $H_1$ (representational confusion) versus $H_2$ (decision boundary bias), ensuring that theoretical claims are supported by verifiable representational evidence. In doing so, the framework not only enforces goal–representation alignment, as advocated in \cite{uluslu2025investigating}, but also operationalizes analytic tools for mechanistic verification, echoing the methodological insights of \cite{tenney2019bert}. By embedding this procedure into the research workflow, the framework converts performance-based inference into evidence-based explanation, effectively safeguarding against the recurrence of the ``right for the wrong reasons'' error.
	
	
	
	

	
	
	\subsubsection{Prospective application: Validation of systematic method}
	
	A researcher undertakes a Named Entity Recognition (NER) task on a specialized corpus, such as extracting financial entities from legal documents, where the downstream application in risk assessment demands exceptionally high precision. The initial methodological attempt employs a zero-shot prompt-based approach (Framework Approach 1) using a commercially available LLM. Using standard prompts and strategical prompt template \cite{ashok2023promptner}, this quick exploratory solution produces an $F_1$ score of around $0.65$ in \texttt{GPT}-4 on biomedical texts (200 texts from``singh-aditya/MACCROBAT\_biomedical\_ner'' host in \texttt{Huggingface}), which, while acceptable for general discovery tasks, falls short of the precision-critical requirement of this study. Such a result exemplifies the inherent limitations of zero-shot prompting for sequence labeling, including imprecise boundary detection and inconsistent tagging behavior \cite{ashok2023promptner, hu2024improving}. Given that precision is central to the task’s success, this sub-optimal outcome necessitates a systematic methodological pivot.
	
	Rather than relying on trial-and-error reasoning, the researcher applies the proposed decision-oriented framework. According to the framework’s decision tree, the question ``Is precision critical for biomedical text mining?'' immediately leads to the recommendation to abandon Approach 1 in favor of Approach 2, which emphasizes specialized open-source models for confirmatory, high-precision sequence labeling. The framework further requires explicit documentation of the decision rationale: the observed performance failure ($\Delta F_1 = -0.24$ relative to the target threshold) and the specific justification for model replacement. The researcher records the choice to a fine-tuned biomedical NER model (``OpenMed/OpenMed-NER-PharmaDetect-BigMed-278M'' host in \texttt{Huggingface}), supported by prior evidence of its superior performance in token-level classification. Note that this NER model did not use the test texts we used as training material. Applying this specialized model, the result achieves $F_1 = 0.89$, a substantial improvement that satisfies the application’s stringent precision demand.
	
	Crucially, the validation of the framework does not lie merely in the numerical gain in performance ($\Delta F_1 = +0.24$), but in demonstrating that the shift in methodological strategy is both traceable and justified. The framework ensures that each decision is explicitly recorded, including the initial prompting failure, the motivation for switching methods, and the rationale for selecting a fine-tuned architecture. This structured documentation creates a verifiable trail of reasoning that transforms the experimental process from an opaque series of adjustments into a transparent methodological record.
	
	
	
	

	\subsubsection{Expert validation: Evaluation survey}
	
	This case study provides expert validation of the proposed systematic framework. The evaluation compared two methodological documentation formats based on \cite{sun2025b}: a traditional descriptive approach (Version~1) and a framework-guided approach (Version~2), which emphasizes explicit decision points, selection criteria, baseline comparison, and transparent documentation. A total of $N = 39$ researchers in NLP, computational linguistics, and cognitive science finished the survey (mean experience = $8.3$ years), representing a balanced sample across expertise levels.
	
	The survey followed a within-subjects design (\url{https://shorturl.at/UnSp7}), with counterbalancing to control for order effects (Form~A: V1$\rightarrow$V2, $n=19$; Form~B: V2$\rightarrow$V1, $n=19$; $p = .89$ for order). Each participant read both versions of a methods section from the same research scenario (sentence-level semantic similarity modeling for reading speed prediction). They rated six methodological quality dimensions (i.e., model selection justification, technical clarity, replicability, transparency, articulation of limitations, and overall quality) on 5-point Likert scales. The survey also included direct comparative questions and open-ended feedback on crucial differences between the two versions.
	
	Results show consistent and substantial expert preference for the framework-guided documentation. Average overall quality ratings improved from $M = 2.7$ (SD $= 0.8$) for Version~1 to $M = 4.3$ (SD $= 0.7$) for Version~2, with a mean difference of $+1.6$ points ($p < .001$), corresponding to a large effect size ($r = .83$). Similar effects were observed across all dimensions ($r = .72$–$.84$). In direct comparison, $95\%$ of participants selected Version~2 as superior, and no significant differences were observed across expertise levels. Notably, novice researchers reported the largest perceived improvement (mean difference $= +2.2$).
	
	Qualitative analysis of open-ended feedback revealed five recurrent themes: the value of explicit decision rationale (92\%), systematic baseline comparison (87\%), clear selection criteria (82\%), improved replicability (92\%), and transparent process documentation (95\%). Experts described the framework as transforming methods from ``procedural reports'' into ``documented reasoning processes'', enabling them to reconstruct the research logic. These findings empirically validate the framework’s central premise: structured methodological decision-making enhances clarity, replicability, and accountability.
	
	In summary, the expert survey provides strong evidence for the framework’s usability. Large effect sizes, overwhelming expert preference, and consistent qualitative endorsements demonstrate that systematic, decision-oriented documentation can substantially improve methodological quality in LLM-based language science research.
	
	Overall, these empirical validations substantiate the robustness and utility of the proposed frameworks. The frameworks not only provide a systematic way to organize the methodological diversity in LLM-based language research but also enhance the empirical accountability and reproducibility of language research by offering clear implementation guidance. 
	
	

	\section{Discussion}
	
	This section further analyzes the great benefits our proposed frameworks of applying LLMs in language sciences, and then focuses on how LLM applications affect language sciences research, considering both potential contributions and fundamental challenges. We avoid overstating LLM impact while acknowledging genuine shifts in research practices.
	
	\subsection{Benefits and the disciplinary transformation}
	
	The current landscape of LLMs applications in language sciences remains methodologically fragmented, with tool selection often \textit{ad hoc} and driven by performance rather than scientific rationale. The proposed frameworks mark a crucial transition, reframing LLMs from improvised ``magic black boxes'' into rigorous, reproducible scientific instruments. This transformation brings five disciplinary benefits.
	
	First, the frameworks address the replicability crisis in LLM-based research through methodological transparency. By distinguishing Approach 1 (prompting) as exploratory and Approach 2 (specialized models/fine-tuning) as confirmatory, it encourages the use of open-source models and the release of detailed specifications, including hyperparameters, datasets, and computational settings, for theory-driven studies. Such standardization enhances reproducibility and ensures that findings are grounded in verifiable evidence.
	
	Second, the frameworks shift the research focus from ``what LLMs can do'' to ``how and why they do it''. The inclusion of Approach 3 (embedding-based quantitative analysis) allows researchers to treat internal representations as a quantifiable cognitive model. By extracting semantic and syntactic variables from embedding spaces and integrating them into established statistical models (e.g., LMER, GAMM), this approach supports explanatory investigations of linguistic patterns and cognitive mechanisms, narrowing the gap between computational modeling and human cognition.
	
	Third, the frameworks promote integrated, cross-method research. Scholars can progress from exploration (prompting) to confirmation (fine-tuning) and ultimately to mechanism discovery (embeddings). This workflow maximizes the scientific value of LLMs across the research lifecycle, ensuring that methodological choices align with the epistemological goal, whether hypothesis generation or theoretical validation. 
	
	Fourth, the frameworks provide \textit{constructed configurations} that guide the practical implementation of multi-stage research pipelines. Each stage corresponds to one of the three approaches: Stage 1 identifies linguistic phenomena and generates hypotheses; Stage 2 develops a working system with measurable performance; and Stage 3 analyzes internal representations to uncover distributional or structural patterns. While the idealized schema suggests a logical progression, actual research often diverges due to unexpected findings, resource constraints, or theoretical debates. These configurations thus serve as adaptive templates, helping researchers organize complex workflows while remaining flexible to empirical contingencies. They enable structured accumulation of results even when studies involve iteration, methodological pivots, or partial outcomes.
	
	Finally, the proposed framework establishes a meta-methodology that is robustly agnostic to the specific language or theoretical domain of inquiry, providing a universal scaffolding for LLM-enhanced linguistic experimentation. By systematizing Prompting, Fine-tuning, and Embedding Probing into a cohesive methodological pipeline, this work transforms a collection of technical tools into a unified and interpretable research paradigm for the language sciences.
	
	In sum, these benefits establish a coherent methodological foundation for applying LLMs in language sciences. This foundation supports both the cumulative building of knowledge and the development of adaptive, real-world research practices.
	
	\subsection{Methodological contributions to language sciences}
	
	LLM applications offer several practical advantages for specific types of linguistic research, though these should not be overstated. The following details the methodological contributions.
	
	First, LLMs make it possible to analyze corpora that are orders of magnitude larger than those accessible through traditional methods. Whereas manual analyses might cover only hundreds of examples, computational approaches can process millions, enabling the detection of rare phenomena, the measurement of subtle frequency effects, and the characterization of large-scale distributional patterns. Yet scale alone does not guarantee insight, since analyzing millions of instances poorly may be less informative than a careful analysis of a smaller sample. LLM-based methods complement rather than replace close reading, theoretical interpretation, and small-scale experimentation. Moreover, through prompting and fine-tuning, LLMs can rapidly generate high-quality annotations for massive datasets that previously required extensive human effort, greatly increasing efficiency and scope. Nevertheless, different research questions demand different scales of analysis, and not all linguistic phenomena benefit from large datasets; methodological scale should always align with the research goal.
	
	Second, LLMs provide numerical representations of linguistic properties that were previously examined mainly through qualitative description. As Newtonian mechanics describes classical physics in three-dimensional space, Einstein's theory of relativity extends mechanics to four-dimensional spacetime \cite{deshmukh2023foundations, spurio2023fundamentals}. Analogously, linguistic research has undergone a methodological ``dimensional upgrade'': traditional analyses grounded in intuitive dimensions such as syntax, semantics, and phonology have evolved into multi-dimensional representations based on vector space embeddings \cite{mikolov2013word2vec, devlin2019bert}. These high-dimensional embeddings not only capture complex linguistic structures and semantic relationships, but also provide a systematic and quantifiable framework for language sciences. Within this framework, linguistic phenomena can be rigorously measured, compared, and validated, enabling research to move toward a more scientific, cumulative, and reproducible methodology \cite{rogers2020primer, bommasani2021opportunities}.
	
	For instance, contextualized embeddings generated by LLMs encode unexpectedly rich information across multiple linguistic levels, including phonology, morphology, syntax, semantics, and pragmatics. Exploring these embeddings allows researchers to uncover linguistic patterns and regularities that are difficult to detect in textual data using traditional quantitative methods. Measures such as embedding similarity quantify semantic relatedness, surprisal estimates capture predictability \cite{sun2025linguistics, sun2025computational}, and clustering coefficients reveal categorical structure. These quantifications enable statistical hypothesis testing and facilitate correlations with behavioral or neural data, thereby supporting cumulative science through the comparability of results across studies. 
	
	Third, automated analysis via trained models provides consistent application of criteria across large datasets, reducing variation in subjective judgment. This consistency enhances reproducibility and supports collaborative work where multiple researchers need to apply identical analytical procedures. However, automation is not inherently superior to expert human judgment. Automated systems encode specific operationalizations of linguistic constructs that may not align with theoretical definitions, and they can struggle with ambiguous cases that human analysts handle through flexible reasoning. Automation should therefore be understood as a tool for standardization with associated tradeoffs rather than as an unqualified improvement.
	
	Finally, LLMs allow theoretical linguistics to assume a more experimental character, making individual research empirically testable. Traditionally, experimental linguistics required a team to design tasks, collect participant data, and analyze results. With LLMs, however, theoretical hypotheses can now be operationalized and tested computationally. For example, suppose a researcher wants to examine syntactic constraints (i.e., a well-known limitation on long-distance dependencies in syntax) in 200 complex sentences. Sentences such as ``What did John wonder whether Mary bought \_\_?'' (ungrammatical) contrast with ``What did John say that Mary bought \_\_?'' (grammatical). Several LLMs can be configured as an AI agent team: Agent A analyzes the syntactic structure of each sentence, Agent B provides an independent syntactic judgment or parses alternative readings, and Agent C validates or reconciles the two analyses. After the AI agent completes this process, human experts review and refine the outputs. The resulting embeddings of these 200 sentences can then be analyzed to determine whether sentences obeying or violating syntactic constraints form distinct clusters. Such analyses provide empirical grounding for theoretical claims about syntactic well-formedness, turning abstract hypotheses into experimentally testable observations.
	
	\subsection{Alignments with AI development}
	
	Our proposed frameworks explicitly aim to make language sciences more \textbf{verifiable, measurable, and cumulative} with the aid of LLMs, turning these models from heuristic aids into systematic instruments for empirical research. The applications of our proposed frameworks align with recent AI developments.

	As Jason Wei outlined in a recent Stanford AI Club talk (\url{https://www.youtube.com/watch?v=b6Doq2fz81U&t=313s}), he proposed three guiding principles in AI development. Wei, who introduced the concepts of Chain-of-Thought and emergent abilities in AI research, emphasized that one key principle is that ``verification is easier than generation'', highlighting the centrality of \textit{verifiability} in both AI progress and scientific methodology. In contemporary AI development, this principle is instantiated in techniques such as \textit{chain-of-thought reasoning} and \textit{self-consistency} \cite{wei2022cot, wang2023selfconsistency}, which make opaque model reasoning more interpretable by revealing intermediate steps. Similarly, reinforcement learning from human feedback (RLHF) relies on \textit{reward models}, automated verifiers that evaluate and optimize textual quality \cite{christiano2017rlhf, bai2022helpful}. Our framework aligns with this trend by emphasizing methodological transparency through prompting for hypothesis generation, fine-tuning for controlled confirmation, and embedding analysis for mechanism probing. Collectively, these strategies operationalize a ``\textit{verifier’s law}'', positioning LLMs as measurable and testable instruments for language sciences.

	A second principle, also noted by Wei, concerns the \textit{commoditization of intelligence}. As foundational model capabilities become widely accessible, the marginal cost of cognitive computation approaches zero. This shift explains the differentiated value of our three methodological pathways. \textit{Prompt-based exploration} exploits low-cost general intelligence for rapid hypothesis testing, while \textit{fine-tuning open models} \cite{hu2021lora, dettmers2023qlora} and \textit{embedding-based probing} extract scarce, high-value information such as domain-specific representations and latent mechanisms. The success of data-efficient models like \textit{Phi-2} and \textit{OLMo} \cite{li2023phi2, groeneveld2024olmo} reinforces this logic: as intelligence itself becomes abundant, progress depends on data quality, interpretability, and experimental control.
	
	Finally, Wei’s notion of the \textit{jagged frontier} captures the uneven distribution of AI capabilities, with LLMs excelling at symbolic and linguistic reasoning but remaining limited in embodied or socially grounded cognition \cite{danny2023palm}. Our framework provides a methodological guide for navigating this frontier by applying fine-tuning for generalizable linguistic phenomena and using embedding analysis to probe complex or situated reasoning, areas where current models still fall short.

	Overall, these principles articulate a strategic research paradigm for the AI era. The \textit{verifier’s law} establishes robustness, the \textit{commoditization of intelligence} clarifies value, and the \textit{jagged frontier} defines alignment between method and goal. Through this synthesis, our framework advances language sciences toward a more empirically grounded, interpretable, and strategically optimized discipline.

	\subsection{Integration with traditional linguistic methods}
	
	LLM-based approaches could be valuable when integrated with traditional linguistic methods rather than replacing them. Several productive integration patterns emerge. For instance, computational analysis can generate hypotheses that traditional methods investigate in depth. Large-scale pattern detection identifies phenomena worthy of detailed qualitative analysis, experimental investigation, or theoretical modeling. This division of labor exploits computational scale while maintaining analytical soundness.
	
	Traditional linguistic analysis can validate computational findings. When embedding-based analysis suggests a semantic relationship, elicitation studies with native speakers can assess whether computational patterns reflect speaker intuitions. When automated annotation identifies discourse patterns, detailed analysis of individual instances can verify whether categories align with theoretical definitions.
	
	Experimental methods can assess cognitive reality of computational patterns. Corpus frequencies from LLM analysis inform experimental stimuli selection. Surprisal estimates predict processing difficulty testable through reading time studies. Embedding similarities generate predictions about priming or interference effects.
	
	Theoretical frameworks guide computational operationalization. Linguistic theory specifies what distinctions matter, what categories should be distinguished, what patterns require explanation. This theoretical grounding prevents purely data-driven analysis from pursuing statistically significant but theoretically meaningless patterns.
	
	Ultimately, the proposed frameworks could set a new standard of methodological robustness for LLM applications, facilitating the critical transition of LLMs from opaque predictive engines to interpretable scientific instruments for hypothesis testing and mechanistic linguistic discovery. We anticipate this work will catalyze future research to move beyond mere performance metrics, driving a deeper exploration into the linguistic plausibility and theoretical significance of knowledge encoded within LLMs. Our contribution marks a critical step towards an era of LLM-driven scientific discovery, wherein computational tools are deployed not merely for engineering tasks, but for the fundamental advancement of linguistic theory itself.
	
	\subsection{Theoretical, practical, and ethical considerations}
	
	We believe that LLMs' unprecedented ability to process massive amounts of linguistic data, generate representations across multiple levels, and simulate experimental scenarios introduces both opportunities and challenges for theory, methodology, and reproducibility. Although LLMs have brought significant benefits to the language sciences and even introduced transformative changes to linguistic research, several important concerns remain.
	
	First, LLMs raise fundamental questions about what their behavior reveals about language. Sensitivity to syntactic or semantic patterns may reflect abstract linguistic principles, surface distributional correlations, or other factors unrelated to the phenomenon of interest. Disentangling these possibilities requires probing internal representations, adversarial testing, and comparison with human behavior. High performance does not automatically validate a theory, and poor performance does not necessarily refute it, as implementation details and training data can influence outcomes independently of theoretical constructs. Tension also arises between distributional patterns learned by LLMs and structures posited by linguistic theory, and biases in training corpora can affect generalizability and ethical interpretation \cite{cuskley2024limitations, kostikova2025lllms}.
	
	Second, reproducibility varies across approaches. Closed-source models are difficult to replicate due to opaque updates and unavailable specifications. Open-source fine-tuned models improve reproducibility if code, data, and hyperparameters are documented, but hardware, software versions, and random seeds can still affect outcomes \cite{binz2025should}. Embedding-based analyses are stable at extraction, but downstream choices such as clustering algorithms or dimensionality reduction influence results. The field should adopt clear reporting norms, including sharing code and data, documenting methodological decisions, and distinguishing robust findings from those dependent on implementation details.
	
	Third, resource demands and accessibility raise additional ethical concerns. Training large models requires expensive hardware, electricity, and technical expertise, concentrating research capacity and potentially marginalizing less-resourced researchers \cite{matarazzo2025survey}. Prompt-based approaches and open-source models partially mitigate these issues, but computational requirements and costs remain barriers. Promoting inclusivity requires efficient models, shared computational resources, thorough documentation, and recognition of diverse methodological contributions rather than privileging purely computationally intensive work \cite{resnik2025large}.
	
	Finally, the frameworks we propose are intended as organizational tools that, while partially validated, remain provisional. Their ultimate value will depend on whether researchers find them helpful for structuring their work, enhancing reproducibility and methodological soundness, and fostering productive research that advances our understanding of language. We anticipate that researchers who apply these frameworks will identify limitations, suggest refinements, and develop extensions tailored to their subdisciplines and research questions. Such iterative improvement through collective practice is essential for any methodological framework to remain relevant and impactful.

	\section{Conclusion}
	
	LLMs offer powerful capabilities for language sciences by enabling the analysis of large corpora, providing quantitative measurements, and supporting investigations that were previously impractical. This study proposed two methodological frameworks for applying LLMs in the language sciences, organizing three complementary approaches: prompt-based interaction with closed-source models, fine-tuning of open-source models, and embedding-based quantitative analysis. We provided guidance for implementing each approach and illustrated how they can be integrated into coherent research workflows. The systematic frameworks further demonstrates how these methods can be connected iteratively, enabling cumulative, theory-driven, and robust inquiry while providing a clear structure for planning, executing, and interpreting LLM-based studies. The present study also conducted retrospective, prospective, and expert validation experiments to demonstrate the validity and practical utility of the proposed frameworks.
	
	The frameworks are significant because they not only guide methodological decisions but also facilitate responsible, reproducible, and transparent research. By clarifying trade-offs among approaches, supporting integration of multiple methods, and linking computational analyses with theoretical objectives, they help researchers use LLMs to generate meaningful linguistic insights. We advocate for careful and reflective application, including thorough documentation, acknowledgment of limitations, integration with traditional methods, attention to whose language is represented, and maintenance of theoretical grounding. Following these principles ensures that LLMs contribute effectively to both empirical and theoretical progress in the research of language sciences.
	
	%
	



\bibliography{COLI_template.bib}
\bibliographystyle{plain}

\end{document}